\documentclass{article}

\PassOptionsToPackage{numbers}{natbib} % round
\usepackage[final]{neurips_2021} % [preprint], [final]

\usepackage[utf8]{inputenc}
\usepackage[T1]{fontenc}
\usepackage{hyperref}
\usepackage{url}
\usepackage{booktabs}
\usepackage{amsfonts}
\usepackage{amsmath}
\usepackage{amsthm}
\usepackage{bm}
\usepackage{nicefrac}
\usepackage{microtype}
\usepackage{graphicx}
\usepackage{subcaption}
\usepackage{xspace}
\usepackage{wrapfig}
\usepackage{tikz}
\usepackage{pifont}
\usepackage{tabularx}
\usepackage{multirow}
\usepackage{xcolor}
\usepackage{tikz}
\usepackage{enumitem}

\hypersetup{
  colorlinks,
  linkcolor={red!50!black},
  citecolor={blue!50!black},
  urlcolor={blue!80!black}
}

\definecolor{hlcolor}{RGB}{206, 225, 237}
\newcommand{\hl}[1]{\smash{\colorbox{hlcolor}{#1}}}
\newtheorem{theorem}{Theorem}
\newtheorem{definition}{Definition}

\newcommand{\Invertible}{{\phi}}
\newcommand{\Hypernet}{{\varphi}}
\newcommand{\Lip}{{\mathrm{Lip}}}
\newcommand{\odesolve}{{\textrm{ODESolve}}}
\newcommand*\diff{\mathop{}\!\mathrm{d}}

\def\Figref#1{Figure~\ref{#1}}

\def\Secref#1{Section~\ref{#1}}
\def\eqref#1{Eq.\ \ref{#1}}
\def\Eqref#1{Equation~\ref{#1}}
\def\1{\bm{1}}

\def\vzero{{\bm{0}}}
\def\vone{{\bm{1}}}
\def\vmu{{\bm{\mu}}}
\def\vsigma{{\bm{\sigma}}}

\def\valpha{{\bm{\alpha}}}
\def\vepsilon{{\bm{\epsilon}}}
\def\vbeta{{\bm{\beta}}}

\def\vc{{\bm{c}}}

\def\vh{{\bm{h}}}

\def\vr{{\bm{r}}}

\def\vt{{\bm{t}}}

\def\vx{{\bm{x}}}

\def\vz{{\bm{z}}}

\def\mA{{\bm{A}}}

\def\mI{{\bm{I}}}

\def\mK{{\bm{K}}}

\def\mQ{{\bm{Q}}}

\def\mV{{\bm{V}}}

\def\mX{{\bm{X}}}

\DeclareMathAlphabet{\mathsfit}{\encodingdefault}{\sfdefault}{m}{sl}
\SetMathAlphabet{\mathsfit}{bold}{\encodingdefault}{\sfdefault}{bx}{n}

\newcommand{\E}{\mathbb{E}}

\newcommand{\R}{\mathbb{R}}

\newcommand{\softmax}{\mathrm{softmax}}
\newcommand{\sigmoid}{\sigma}

\DeclareMathOperator{\sign}{sign}

\title{Neural Flows: Efficient Alternative to Neural ODEs}

\author{
  Marin Bilo\v{s}$^1$\thanks{Work partially done during an internship at Amazon Research. Correspondence to: \texttt{bilos@in.tum.de}.},
  Johanna Sommer$^1$,
  Syama Sundar Rangapuram$^2$,\\
  \textbf{
  Tim Januschowski$^2$,
  Stephan G\"unnemann$^1$
  } \\
  $^1$Technical University of Munich, $^2$AWS AI Labs, Germany
}

\begin{document}

\maketitle

\begin{abstract}
Neural ordinary differential equations describe how values change in time.
This is the reason why they gained importance in modeling sequential data, especially when the observations are made at irregular intervals.
In this paper we propose an alternative by directly modeling the solution curves --- the flow of an ODE --- with a neural network.
This immediately eliminates the need for expensive numerical solvers while still maintaining the modeling capability of neural ODEs.
We propose several flow architectures suitable for different applications by establishing precise conditions on when a function defines a valid flow.
Apart from computational efficiency, we also provide empirical evidence of favorable generalization performance via applications in time series modeling, forecasting, and density estimation.
\end{abstract}

%!TEX root = ../main.tex

\section{Introduction}

Ordinary differential equations (ODEs) are among the most important tools for modeling complex systems, both in natural and social sciences.
They describe the \textit{instantaneous change} in the system, which is often an easier way to model physical phenomena than specifying the whole system itself.
For example, the change of the pendulum angle or the change in population can be naturally expressed in the differential form. Similarly, \citet{chen2018neural} introduce neural ODEs that describe how some quantity of interest represented as a vector $\vx$, changes with time: $\dot{\vx} = f(t, \vx(t))$, where $f$ is now a neural network. Starting at some initial value $\vx(t_0)$ we can find the result of this dynamic at any $t_1$:
\begin{align}\label{eq:ode_solution}
    \vx(t_1) = \vx(t_0) + \int_{t_0}^{t_1} f(t, \vx(t)) \diff t = \odesolve(\vx(t_0), f, t_0, t_1).
\end{align}
\begin{wrapfigure}[16]{r}{0.33\textwidth}
    \vspace*{-0.7cm}
    \centering
    \begin{tikzpicture}
        \node () at (0,0) {\input{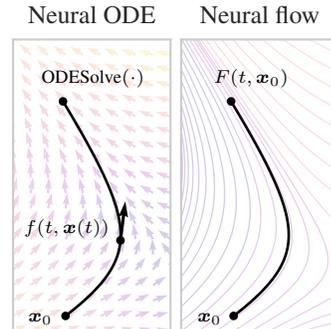}};
        \node () at (-1.8,-2) {\scriptsize $\vx_0$};
        \node () at (-1.1,1.2) {\scriptsize $\odesolve(\cdot)$};
        \node () at (-1.45,-0.8) {\scriptsize $f(t, \vx(t))$};
        \node () at (0.45,-2) {\scriptsize $\vx_0$};
        \node () at (1.0,1.2) {\scriptsize $F(t, \vx_0)$};
    \end{tikzpicture}
    \vspace{-0.5cm}
    \caption{(Left) ODE requires numerical solver which evaluates $f$ at many points along the solution curve. (Right) Our approach returns the solutions directly.}
    \label{fig:fig1}
\end{wrapfigure}
It is sufficient for $f$ to be continuous in $t$ and Lipschitz continuous in $\vx$ to have a unique solution, by the Picard–Lindel\"of theorem \citep{coddington1955theory}. This mild condition is already satisfied by a large family of neural networks. In most practically relevant scenarios, the integral in \Eqref{eq:ode_solution} has to be solved numerically, requiring a trade-off between computation cost and numerical precision. Much of the follow up work to \citep{chen2018neural} focused on retaining expressive dynamics while requiring fewer solver evaluations \citep[][]{finlay2020train,kelly2020learning}.

In the machine learning context we are given a set of initial conditions (often at $t_0=0$) and a loss function for the solution evaluated at time $t_1$. One example is modeling time series where the latent state is evolved in continuous time and is used to predict the observed measurements \citep{de2019gru}. Here, unlike in physics for example, the function $f$ is completely unknown and needs to be learned from data. Thus, \citep{chen2018neural} used neural networks to model it, for their ability to capture complex dynamics. However, note that this comes at the cost of the ODE being non-interpretable.

Since solving an ODE is expensive, we want to find a way to keep the desired properties of neural ODEs at a much smaller computation cost.
If we take a step back, we see that neural ODEs take initial values as inputs and return non-intersecting solution curves (\Figref{fig:fig1}). In this paper we propose to model the solution curves directly, with a neural network, instead of specifying the derivative. That is, given an initial condition we return the solution with a single forward pass through our network. Straight away, this leads to improvements in computation performance because we avoid using ODE solvers altogether. We show how our method can be used as a faster alternative to ODEs in existing models \citep{chen2021spatio,de2019gru,jia2019jump,rubanova2019latent}, while improving the modeling performance. In the following, we derive the conditions that our method needs to satisfy and propose different architectures that implement them.

%!TEX root = ../main.tex

\section{Neural flows}\label{sec:neural_flows}

In this section, we present our method, \textit{neural flows}, that directly models the solution curve of an ODE with a neural network. For simplicity, let us briefly assume that the initial condition $\vx_0 = \vx(t_0)$ is specified at $t_0 = 0$. We handle the general case shortly. Then, \Eqref{eq:ode_solution} can be written as $\vx(t) = F(t, \vx_0)$, where $F$ is the solution to the initial value problem, $\dot{\vx} = f(t, \vx(t)), \vx_0 = \vx(0)$. We will model $F$ with a neural network. For this, we first list the conditions that $F$ must satisfy so that it is a solution to some ODE. Let $F : [0, T] \times \R^d \rightarrow \R^d$ be a smooth function satisfying:
\renewcommand{\labelenumi}{\roman{enumi})}
\begin{enumerate}
    \item $F(0, \vx_0) = \vx_0$, (initial condition)
    \item $F(t, \cdot)$ is invertible, $\forall t$. (uniqueness of the solution given the initial value $\vx_0$; i.e., the curves specified by $F$ corresponding to different initial values should not intersect for any $t$)
\end{enumerate}
There is an exact correspondence between a function $F$ with the above properties and an ODE defined with $f$ such that the derivative $\frac{\diff}{\diff t} F(t, \vx_0) $ matches $f(t, \vx(t))$ everywhere, given $\vx_0 = \vx(0)$ \citep[Theorem 9.12]{lee2013smooth}.
In general, we can say that $f$ defines a vector field and $F$ defines a family of integral curves, also known as the \textit{flow} in mathematics (not to be confused with normalizing flow).
As $F$ will be parameterized with a neural network, condition i) requires that its parameters must depend on $t$ such that we have the identity map at $t=0$.

Note that by providing $\vx_0$ we define a smooth trajectory $F(\cdot, \vx_0)$ --- the solution to some ODE with the initial condition at $t_0=0$. If we relax the restriction $t_0 = 0$ and allow $\vx_0$ to be specified at an arbitrary $t_0 \in \R$, the solution can be obtained with a simple procedure.
We first go back to the case $t=0$ where we obtain the corresponding ``initial'' value $\hat{\vx}_0 := \vx(0) = F^{-1}(t_0, \vx_0)$.
This then gives us the required solution $F(\cdot, \hat{\vx}_0)$ to the original initial value problem. Thus, we often prefer functions with an analytical inverse.

Finally, we tackle implementing $F$. The second property instructs us that the function $F(t, \cdot)$ is a diffeomorphism on $\R^d$. We can satisfy this by drawing inspiration from existing works on normalizing flows and invertible neural networks \citep[e.g.,][]{dinh2016density,behrmann2019invertible}. In our case, the parameters must be conditioned on time, with identity at $t=0$.
As a starting example, consider a linear ODE $f(t, \vx(t)) = \mA \vx(t)$, with $\vx(0) = \vx_0$. Its solution can be expressed as $F(t, \vx_0) = \exp(\mA t) \vx_0$, where $\exp$ is the matrix exponential. Here, the learnable parameters $\mA$ are simply multiplied by $t$ to ensure condition i); and given fixed $t$, the network behaves as an invertible linear transformation. In the following we propose other, more expressive functions suitable for applications such as time series modeling.

\subsection{Proposed flow architectures}\label{sec:proposed_architectures}
\textbf{ResNet flow.}
A single residual layer $\vx_{t+1} = \vx_t + g(\vx_t)$ \citep{he2016deep} bears a resemblance to \Eqref{eq:ode_solution} and can be seen as a discretized version of a continuous transformation which inspired the development of neural ODEs.
Although plain ResNets are not invertible, one could use spectral normalization~\citep{gouk2021regularisation} to enforce a small Lipschitz constant of the network, which guarantees invertibility \citep[Theorem 1]{behrmann2019invertible}.
Thus, ResNets become a natural choice for modeling the solution curve $F$ resulting in the following extension --- ResNet flow:
\begin{align}\label{eq:resnet_flow}
    F(t, \vx) = \vx + \Hypernet(t) g(t, \vx),
\end{align}
where $\Hypernet:\R \rightarrow \R^d$. This satisfies properties i) and ii) from above when $\Hypernet(0) = \vzero$ and $|\Hypernet(t)_i| < 1$; and $g : \R^{d+1} \rightarrow \R^d$ is an arbitrary contractive neural network ($\Lip(g) < 1$). One simple choice for $\Hypernet$ is a $\tanh$ function.
The inverse of $F$ can be found via fixed point iteration similar to~\citep{behrmann2019invertible}.

\textbf{GRU flow.}
Time series data is traditionally modeled with recurrent neural networks, e.g., with a GRU \citep{cho2014gru}, such that the hidden state $\vh_{t-1}$ is updated at fixed intervals with the new observation $\vx_t$:
\begin{align}\label{eq:gru}
\begin{aligned}
    \vh_{t} &= \textrm{GRUCell}(\vh_{t-1}, \vx_t) = \vz_t \odot \vh_{t-1} + (1 - \vz_t) \odot \vc_t,
\end{aligned}
\end{align}
where $\vz_t$ and $\vc_t$ are functions of the previous state $\vh_{t-1}$ and the new input $\vx_t$.

\citet{de2019gru} derived the continuous equivalent of this architecture called GRU-ODE (see Appendix~\ref{app:gru_ode}). Given the initial condition $\vh_{0} = \vh(t_0)$, they evolve the hidden state $\vh(t)$ with an ODE, until they observe new $\vx_{t_1}$ at time $t_1$, when they use \Eqref{eq:gru} to update it:
\begin{align}\label{eq:gru-ode}
    \bar{\vh}_{t_1} = \odesolve(\vh_0, \textrm{GRU-ODE}, t_0, t_1),
    \quad \vh_{t_1} = \textrm{GRUCell}(\bar{\vh}_{t_1}, \vx_{t_1}).
\end{align}
Here, we will derive the flow version of GRU-ODE. If we rewrite \Eqref{eq:gru} by regrouping terms: $\vh_{t} = \vh_{t-1} + (1 - \vz_t) \odot (\vc_t - \vh_{t-1})$, we see that GRU update acts as a single ResNet layer.
\begin{definition}\label{def:gru_flow}
Let $f_z, f_r, f_c : \R^{d+1}\rightarrow \R^d$ be any arbitrary neural networks and let $z(t, \vh) = \alpha \cdot \sigma(f_z(t, \vh))$, $r(t, \vh) = \beta \cdot \sigma(f_r(t, \vh))$, $c(t, \vh) = \tanh(f_c(t, r(t, \vh) \odot \vh))$, where $\alpha, \beta \in \R$ and $\sigmoid$ is a sigmoid function. Further, let $\Hypernet : \R \rightarrow \R^d$ be a continuous function with $\Hypernet(0) = \vzero$ and $|\Hypernet(t)_i| < 1$.
Then the evolution of GRU state in continuous time is defined as:
\begin{align}\label{eq:gru_continuous}
	F(t, \vh) = \vh + \Hypernet(t) (1 - z(t, \vh)) \odot (c(t, \vh) - \vh).
\end{align}
\end{definition}
\begin{theorem}\label{thm:gru_invertible}
    A neural network defined by \Eqref{eq:gru_continuous} specifies a flow when the functions $f_z$, $f_r$ and $f_c$ are contractive maps, i.e., $\Lip(f_{\cdot}) < 1$, and $\alpha = \frac{2}{5}$, $\beta = \frac{4}{5}$.
\end{theorem}
We prove Theorem~\ref{thm:gru_invertible} in Appendix~\ref{app:gru_flow} by showing that the second summand on the right hand side in \Eqref{eq:gru_continuous} satisfies Lipschitz constraint making the whole network invertible. We also show that the GRU flow has the same desired properties as GRU-ODE, namely, bounding the hidden state in $(-1, 1)$ and having the Lipschitz constant of $2$.
Note that GRU flow (\Eqref{eq:gru_continuous}) acts as a replacement to \odesolve~in \Eqref{eq:gru-ode}. Alternatively, we can append $\vx_t$ to the input of $f_z, f_r$ and $f_c$, which would give us a continuous-in-time version of GRU.

\textbf{Coupling flow.}
The disadvantage of both ResNet flow and GRU flow is the missing analytical inverse.
To this end, we propose a continuous-in-time version of an invertible transformation based on splitting the input dimensions into two disjoint sets $A$ and $B$, $A \cup B = \{1, 2, \dots, d\}$ \citep{dinh2016density}. We copy the values indexed by $B$ and transform the rest conditioned on $\vx_B$ which gives us the coupling flow:
\begin{align}\label{eq:coupling_flow}
    F(t, \vx)_A = \vx_{A} \exp(u(t, \vx_{B}) \Hypernet_u(t)) + v(t, \vx_{B}) \Hypernet_v(t),
\end{align}
where $u$, $v$ are arbitrary neural networks and $\Hypernet_u(0) = \Hypernet_v(0) = \vzero$. We can easily see that this satisfies condition i), and it is invertible by design regardless of $t$ \citep{dinh2016density}. Since some values stay constant in a single layer, we apply multiple consecutive transformations, choosing different partitions $A$ and $B$.

For all three models we can stack multiple layers $F = F_1 \circ \dots \circ F_n$ and still define a proper flow since the composition of invertible functions is invertible, and consecutive identities give an identity.

We can think of $\Hypernet$ (including $\Hypernet_u, \Hypernet_v$) as a time embedding function that has to be zero at $t=0$. Since it is a function of a single variable, we would like to keep the complexity low and avoid using general neural networks in favor of interpretable and expressive basis functions. A simple example is linear dependence on time $\Hypernet(t) = \valpha t$, or $\tanh(\valpha t)$ for ResNet flow. We use these in the experiments. An alternative, more powerful embedding consists of Fourier features $\smash{\Hypernet(t)_i = \sum_k \valpha_{ik} \sin(\vbeta_{ik} t)}$.

\subsection{On approximation capabilities}

Previous works established that neural ODEs are $sup$-universal for diffeomorphic functions \citep{teshima2020universal} and are $L^p$-universal for continuous maps when composed with terminal family \citep{li2019deep}. A similar result also holds for affine coupling flows \citep{teshima2020coupling}, whereas general residual networks can approximate any function \citep{lin2018resnet}. The ResNet flow, as defined in \Eqref{eq:resnet_flow}, can be viewed as an Euler discretization, meaning it is enough to stack appropriately many layers to uniformly approximate any ODE solution \citep{li2019deep}. GRU flow can be viewed as a ResNet flow and coupling flow shares a similar structure, meaning that if we can set them to act as an Euler discretization we can match any ODE. However, this is of limited use in practice since we use finitely many layers, so the main focus of this paper is to provide the empirical evidence that we can outperform neural ODEs on relevant real-world tasks.

Other results \citep{dupont2019augmented,zhang2020approximation} consider limitations of neural ODEs in modeling general homeomorphisms (e.g., $x \mapsto -x$) and propose the solution that adds dimensions to the input $\vx$. Such augmented networks can model higher order dynamics. This can be explicitly defined through certain constraints for further improvements in performance and better interpretability \citep{norcliffe2020second}. We can apply the same trick to our models. However, instead of augmenting $\vx$, a simpler solution is to relax the conditions on $F$ given the task. For example, if we do not need invertibility, we can remove the Lipschitz constraint in \Eqref{eq:resnet_flow}. Since neural flows offer such flexibility, they might be of more practical relevance in these use cases.

%!TEX root = ../main.tex

%%% Irregular Time Series
\section{Applications}\label{sec:applications}

In this section we review two main applications of neural ODEs: modeling irregularly-sampled time series and density estimation. We describe the existing modeling approaches and propose extensions using neural flows. In \Secref{sec:experiments} we will use models presented here to qualitatively and quantitatively compare neural flows with neural ODEs.

\subsection{Continuous-time latent variable models}\label{sec:latent_model}

Autoregressive \citep{oord2016wavenet,salinas2020deepar} and state space models~\citep{Hyndman2008,rangapuram2018deep} have achieved considerable success modeling regularly-sampled time series. However, many real-world applications do not have a constant sampling rate and may contain missing values, e.g., in healthcare we have very sparse measurements at irregular time intervals. Here we describe how our neural flow models can be used in such scenario.

\textbf{Encoder.}
In this setting, we are given a sequence of observations $\mX = (\vx_1, \dots, \vx_n)$, $\vx_i \in \R^d$ at times $\vt = (t_1, \dots, t_n)$. To represent this type of data, previous RNN-based works relied on exponentially decaying hidden state \citep{che2018recurrent}, time gating \citep{neil2016phased}, or simply adding time as an additional input \citep{du2016recurrent}. More recently, various ODE-based models built on top of RNNs to evolve the hidden state between observations in continuous time, giving rise to, e.g., ODE-RNN \citep{rubanova2019latent}, while outperforming previous approaches. Another model is GRU-ODE \citep{de2019gru}, which we already described in \Eqref{eq:gru-ode}. We proposed the GRU flow (\Eqref{eq:gru_continuous}) that can be used as a straightforward replacement.

\citet{lechner2020learning} showed that simply evolving the hidden state with a neural ODE can cause vanishing or exploding gradients, a known issue in RNNs \citep{bengio1994learning}. Thus, they propose using an LSTM-based \citep{hochreiter1997long} model instead.
The difference to ODE-RNN \citep{rubanova2019latent} is using an $\textrm{LSTMCell}$ and introducing another hidden state that is not updated continuously in time, which in turn allows gradient propagation via internal LSTM gating. To adapt this to our framework, we simply replace the ODESolve with the ResNet or coupling flow to obtain a neural flow model.

\textbf{Decoder.} Once we have a hidden state representation $\vh_{i}$ of the irregularly-sampled sequence up to $\vx_i$, we are interested in making future predictions. The ODE based models continue evolving the hidden state using a numerical solver to get the representation at time $t_{i+1}$, with $\vh_{i+1} = \odesolve(\vh_i, f, t_i, t_{i+1})$. With neural flows we can simply pass the next time point $t_{i+1}$ into $F$ and get the next hidden state directly. In the following we show how the presented encoder-decoder model is used in both the smoothing and filtering approaches for irregular time series modeling.

\textbf{Smoothing approach.}
The given sequence of observations $(\mX, \vt)$ is modeled with latent variables or states $(\vz_1, \dots, \vz_n) \sim \R^h$, such that $\vx_i \sim p(\vx_i | \vz_i)$, conditionally independent of other $\vx_j$~\citep{chen2018neural, rubanova2019latent}.
There is a predesignated prior state $\vz_0$ at $t=0$ from which the latent state is assumed to evolve continuously.
More precisely, if $z_{0}$ is a sample from the initial latent state $\vz_0$, then a latent state sample at any future time step $t$ is given by $\smash{z_{t} = F(t, z_0)}$.

Since the exact inference on the initial state $\vz_0$, $p(\vz_0 | \mX, \vt)$,  is intractable, we proceed by doing approximate inference following the variational auto-encoder approach \citep{chen2018neural,rubanova2019latent}.
We use an LSTM-based neural flow encoder that processes $(\mX, \vt)$ and outputs the approximate posterior parameters $\vmu$ and $\vsigma$ from the last state, $q(\vz_0 | \mX, \vt) = \mathcal{N}(\vmu, \vsigma)$. The decoder returns all $z_i$ deterministically at times $\vt$ with $F(t, z_0)$, with initial condition $z_0 \sim q(\vz_0 | \mX, \vt)$.
For the latent state at an arbitrary $t_i$, the target is generated according to the model $\vx_i \sim p(\vx_i | \vz_i)$.
Given $p(\vz_0) = \mathcal{N}(\vzero, \vone)$, the overall model is trained by maximizing the evidence lower bound:
\begin{align}\label{eq:elbo}
    \mathrm{ELBO} = \E_{z_0 \sim q(\vz_0| \mX, \vt))}[\log p(\mX)] - \mathrm{KL}[q(\vz_0 | \mX, \vt) || p(\vz_0)].
\end{align}
Using continuous time models brings up multiple advantages, from handling irregular time points automatically to making predictions at any, and as many time points as required, allowing us to do reconstruction, missing value imputation and forecasting.
This holds whether we use neural flows or ODEs, but our approach is more computationally efficient, which matters as we scale to bigger data.

\textbf{Filtering approach.} In contrast to the previous approach, we can alternatively do the inference in an online fashion at each of the observed time points, i.e., estimating the posterior $p(\vz_i | \vx_{1:i}, \vt_{1:i})$ after seeing observations until the current time step $i$.
This is known as filtering.
Here, the prediction for future time steps is done by evolving the posterior corresponding to the final observed time point $p(\vz_n | \mX, \vt)$ instead of the initial time point $p(\vz_0 | \mX, \vt)$, as was done in the smoothing approach.

In this paper, we follow the general approach suggested by~\citet{de2019gru} for capturing non-linear dynamics.
We use GRU flow (instead of GRU-ODE) for evolving the hidden state $\vh_i \in \R^h$ and we output the mean and variance of the approximate posterior $q(\vz_i | \vx_{1:i}, \vt_{1:i})$.
The log-likelihood cannot be computed exactly under this model so~\citep{de2019gru} suggest using a custom objective that is the analogue to Bayesian filtering (see Appendix~\ref{app:gru_ode_loss} for details). Unlike~\citep{de2019gru}, which needs to solve the ODE for every observation, our method only needs a single pass through the network per observation.

%%% TPP
\subsection{Temporal point processes}\label{sec:tpp}

Sometimes temporal data is measured irregularly \textit{and} the times at which we observe the events come from some underlying process modeled with temporal point processes (TPPs). For example, we can use TPPs to model the times of messages between users. One example type of behavior we want to capture is excitation \citep{hawkes1971spectra}, e.g., observing one message increases the chance of seeing other soon after.

A realization of a TPP on an interval $[0, T]$ is an increasing sequence of arrival times $\vt = (t_1, \dots, t_n)$, $t_i \in [0, T]$, where $n$ is a random variable. The model is defined with an intensity function $\lambda(t)$ that tells us how many events we expect to see in some bounded area \citep{daleyintroduction}. The intensity has to be positive. We define the history $\mathcal{H}_{t_i}$ as the events that precede $t_i$, and further define the conditional intensity function $\lambda^*(t)$ which depends on history. For convenience, we can also work with inter-event times $\tau_i = t_i - t_{i-1}$, without losing generality. We train the model by maximizing the log-likelihood:
\begin{align}\label{eq:tpp_likelihood}
    \log p(\vt) = \sum_i^n \log \lambda^{*}(t_i) - \int_0^T \lambda^{*}(s) \diff s .
\end{align}
Previous works \citep{shchur2021neural} used autoregressive models (e.g., RNNs) to represent the history with a fixed-size vector $\vh_i$ \citep{du2016recurrent}. The intensity function can correspond to a simple distribution \citep{du2016recurrent} or a mixture of distributions \citep{shchur2019intensity}. Then the integral in \Eqref{eq:tpp_likelihood} can be computed exactly. Another possibility is modeling $\lambda(t)$ with an arbitrary neural network which requires Monte Carlo integration \citep{bilos2019uncertainty,mei2017neural}. On the other hand, \citet{jia2019jump} propose a jump ODE model that evolves the hidden state $\vh(t)$ with an ODE and updates the state with new observations, similar to LSTM-ODE. In this case, obtaining the hidden state and solving the integral in \Eqref{eq:tpp_likelihood} can be done in a single solver call.

\textbf{Marked point processes.}
Often, we are also interested in what type of an event happened at time point $t_i$. Thus, we can assign the observed type $\vx_i$, also called mark, and model the arrival times and marks jointly: $p(\vt, \mX) = p(\vt) p(\mX | \vt)$. Written like this, we can keep the model for arrival times as in \Eqref{eq:tpp_likelihood}, and add a module that inputs the history $\vh_i$ and the next time point $t_{i+1}$ and outputs the probabilities for each mark type. The special case of $\vx_i \in \R^d$ is covered in the next section.

\subsection{Time-dependent density estimation}\label{sec:time_density}

Normalizing flows (NFs) define densities with invertible transformations of random variables. That is, given a random variable $\vz \sim q(\vz)$, $\vz \in \R^d$ and an invertible function $F : \R^d \rightarrow \R^d$, we can compute the probability density function of $\vx = F(\vz)$ with the change of variables formula \citep{papamakarios2019normalizing}: $\smash{p(\vx) = q(\vz) | \det J_F(\vz) |^{-1}}$,
where $J_F$ is the Jacobian of $F$. As we can see, it is important to define a function $F$ that is easily invertible and has a tractable determinant of the Jacobian. One example is the coupling NF \citep{dinh2016density}, which we used to construct the coupling flow in \Eqref{eq:coupling_flow}. Other tractable models include autoregressive \citep{kingma2016improved,papamakarios2017masked} and matrix factorization based NFs \citep{berg2018sylvester,kingma2018glow}.

In contrast to this, \citet{chen2018neural} define the transformation with an ODE: $f(t, \vz(t)) = \frac{\partial}{\partial t} \vz(t)$. This allows them to define the instantaneous change in log-density as well as the continuous equivalent to the change of variables formula, giving rise to the continuous normalizing flow (CNF):
\begin{align}\label{eq:cnf}
    \frac{\partial}{\partial t}\log p(\vz(t)) = -\mathrm{tr}\left( \frac{\partial f}{\partial \vz(t)} \right),\quad
    \log p(\vx) = \log q(\vz(t_0)) - \int_{t_0}^{t_1} \mathrm{tr} \left( \frac{\partial f}{\partial \vz(t)} \right) \diff t,
\end{align}
where $t_0 = 0$ and $t_1 = 1$ are usually fixed.
The neural network $f$ can be arbitrary as long as it gives unique ODE solutions. This offers an advantage when we need special structure of $f$ that cannot be easily implemented with the discrete NFs, e.g., in physics we often require equivariant transformations \citep{bilos2021scalable,kohler2020equivariant}. Besides the cost of running the solver, calculating the trace at each step in \Eqref{eq:cnf} becomes intractable as the dimension of data grows, so one resorts to stochastic estimation \citep{grathwohl2018ffjord}. A similar approximation method is used for estimating the determinant in an invertible ResNet model \citep{behrmann2019invertible}. We discuss the computation complexity in Appendix~\ref{app:trace_complexity}. Again, if we consider a linear ODE, we can easily show that calculating the trace and calculating the determinant of the corresponding flow is equivalent (see Appendix~\ref{app:linear_ode_density}).

However, we are not interested in comparison between different normalizing flows for stationary densities \citep[see e.g.,][]{kobyzev2020normalizing}, since \textit{flow endpoints} $t_0$ and $t_1$ are always fixed; thus, our models would be reduced to the discrete NFs. Recently, \citet{chen2021spatio} demonstrated how CNFs can evolve the densities in continuous time, with varying $t_0$ and $t_1$, which proves useful for spatio-temporal data. We will show how to do the same with our coupling flow, something that has not been explored before.

\textbf{Spatio-temporal processes.}
We reuse the notation from \Secref{sec:tpp} to denote the arrival times with $\vt$ and marks with $\mX$, $\vx_i \in \R^d$, which are now continuous variables. Values $\vx_i$ often correspond to locations of events, e.g., earthquakes \citep{ogata1984inference} or disease outbreaks \citep{meyer2012space}. We use the temporal point processes from \Secref{sec:tpp} to model $p(\vt)$, and are only left with the conditional density $p(\mX | \vt)$.
\citet{chen2021spatio} propose several models for this, the first one being the time-varying CNF where $p(\vx_i | t_i)$ is estimated by integrating \Eqref{eq:cnf} from $t_0 = 0$ to observed $t_i$. Using our affine coupling flow as defined in \Eqref{eq:coupling_flow} we can write:
\begin{align}
    p(\vx_i | t_i) = q(F^{-1}(t_i, \vx_i)) | \det J_{F^{-1}}(\vx_i) |,
\end{align}
where $q$ is the base density (defined with any NF) and the determinant is the product of the diagonal values of the Jacobian w.r.t.\ $\vx_i$, which are simply $\exp$ terms from \Eqref{eq:coupling_flow} \citep{dinh2016density}. The density $p$ evolves with time, the same way as in the CNF model, but without using the solver or trace estimation. To generate new realizations at $t_i$, we first sample from $q$ to get $\vx_0 \sim q(\vx_0)$, then evaluate $F(t_i, \vx_0)$.

An alternative model, attentive CNF~\citep{chen2021spatio}, is more expressive compared to the time-varying CNF and more efficient than jump ODE models \citep{chen2021spatio,jia2019jump}. The probability density of $\vx_i$ depends on all the previous values $\vx_{j<i}$ through the attention mechanism \citep{vaswani2017attention}. In our model, we represent all the previous points $\vx_{j<i}$ with an attention encoder and define a conditional coupling NF $p(\vx_i | t_i, \vx_{j<i})$. We describe the full model in Appendix~\ref{app:attentive_nf}. Both of the previous models can also use ResNet flow, but the benefits over ODEs vanish since the determinant and the inverse require iterative procedure.

%!TEX root = ../main.tex

\section{Experiments}\label{sec:experiments}
In this section we show that flow-based models can match or outperform ODEs at a smaller computation cost, both in latent variable time series modeling, as well as TPPs and time-dependent density estimation.
To make fair comparison, we used recently introduced reparameterization trick for ODEs that allows faster mini-batching \citep{chen2021spatio}, and the semi-norm trick for faster backpropagation \citep{kidger2020hey}, making the models more competitive compared to the original works.
In all experiments we split the data into train, validation and test set; train with early stopping and report results on test set. We use Adam optimizer \citep{kingma2014adam}.
% c5.4xlarge and p3.2xlarge
For training we use two different machines, one with 3.4GHz processor and 32GB RAM and another with 61GB RAM and NVIDIA Tesla V100 GPU 16GB \citep{liberty2020elastic}.
All datasets are publicly available, we include the download links and release the code that reproduces the results.\footnote{\url{https://www.daml.in.tum.de/neural-flows}}

\textbf{Synthetic data.}
We compare the performance of neural ODEs and neural flows on periodic signals and data generated from autonomous ODEs. Full setup and results are presented in Appendix~\ref{app:synthetic_experiment}. In short, we observe that training with adaptive solvers \citep{dormand1980family} is slower compared to fixed-step solvers, as expected. With the fixed step, however, we are not guaranteed invertibility \citep{ott2020neural}, which can be an issue in, e.g., density estimation. Using the same setup, our models are an order of magnitude faster. Finally, neural ODEs struggle with non-smooth signals while neural flows perform much better, although they also only output smooth dynamics. Neural flows are also better at extrapolating, although none of the models excel in this task.

\textbf{Stiff ODEs.}
The numerical approach to solving ODEs is not only slow but it can be unstable. This can happen when the ODE becomes stiff, i.e., the solver needs to take very small steps even though the solution curve is smooth. For neural ODEs, it can happen that the target dynamic is known to be stiff or the latent dynamic becomes stiff during training.

\begin{wrapfigure}[8]{r}{0.4\textwidth}
    \centering
    \vspace*{-0.9cm}
    \input{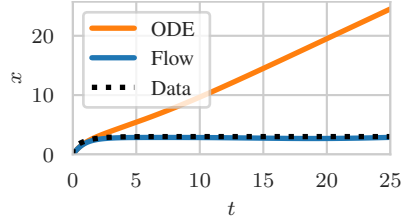}
    \vspace*{-0.2cm}
    \caption{Flows handle stiffness better.}
    \label{fig:stiffness}
\end{wrapfigure}

To see the effects of this, we use the experiment from \citep{ghosh2020steer}. The ODE is given by: $\dot{x} = -1000 x + 3000 - 2000 e^{-t}$. We train a neural ODE model and a coupling flow to match the data, minimizing MSE. The data contains initial conditions and solutions, on small intervals with $t_2 -t_1 = 0.125$, $t\in[0, 15]$. The flow first finds the solution at $t_0=0$ and then solves for $t_2$ (\Secref{sec:neural_flows}). We evaluate on an extended time interval given $x_0 = 0$. \Figref{fig:stiffness} shows that the neural ODE with an adaptive solver does not match the correct solution, due to its stiffness. In contrast, flow captures the solution correctly, as expected, since it does not use a numerical solver.
% The test MSE is lower compared to an ODE with temporal regularization \citep{ghosh2020steer}.

\textbf{Smoothing approach.}
Following \citep{rubanova2019latent}, we use three datasets: \underline{\smash{Activity}}, \underline{\smash{Physionet}}, and \underline{\smash{MuJoCo}}. Activity contains 6554 time series of 3d positions of 4 sensors attached to an individual. The goal is to classify one of the 7 possible activities (e.g., walking, lying, etc.). Physionet \citep{silva2012predicting} contains 8000 time series and 37 features of patients' measurements from the first 48 hours after being admitted to ICU. The goal is to predict the mortality. MuJoCo is created from a simple physics simulation ``Hopper'' \citep{tassa2018deepmind} by randomly sampling initial positions and velocities and letting dynamics evolve deterministically in time. There are 10000 sequences, with 100 time steps and 14 features.

We use the encoder-decoder model (\Secref{sec:latent_model}) and maximize \Eqref{eq:elbo}. We use the same number of hidden layers and the same size of latent states for both the neural ODE, coupling flow and ResNet flow, giving approximately the same number of trainable parameters.
ODE models use either Euler or adaptive solvers and we report the best results.
The results in Table~\ref{tab:encoder_decoder_results} show the reconstruction error and the accuracy of prediction.
For better readability, we scale MSE scores same as in \citep{rubanova2019latent}. % by $10^{2}$ for Activity and by $10^{3}$ for others.
Neural flows outperform ODE models everywhere (Physionet reconstruction within the confidence interval). We noticed that it is possible to further improve the results with bigger flow models but we focused on having similar sized models to show that we can get better results at a much smaller cost.

\textbf{Speed improvements.}
In the smoothing experiment, our method offers more than two times speed-up during training compared to an ODE using an Euler method (\Figref{fig:speed_comparison_all}, different boxes corresponding to different datasets, grouped by experiment types). The gap is even larger for adaptive solvers. Note that \Figref{fig:speed_comparison_all} shows an average time to run one training epoch which includes other operations, such as data fetching, state update etc. This shows that ODESolve contributes significantly to long training times. When comparing ODEs and flows alone, our method is much faster. In the following we will discuss the results from \Figref{fig:speed_comparison_all} for other experiments as well as other results.

\begin{table}
    \centering
    \begin{tabular}{lcccccccc}
                  &          MuJoCo & \multicolumn{2}{c}{Activity} & \multicolumn{2}{c}{Physionet} \\
             &MSE  &MSE &Accuracy &MSE &AUC\\
    \midrule
    Neural ODE    & 8.403{\small $\pm$0.142} & 6.390{\small $\pm$0.136} & \hl{0.756{\small {\small $\pm$0.013}}} & \hl{\textbf{4.833{\small $\pm$0.078}}} & 0.777{\small $\pm$0.012} \\
    Coupling flow & \textbf{\hl{4.217{\small $\pm$0.147}}} & 6.579{\small $\pm$0.049} & 0.752{\small $\pm$0.012} & \hl{4.860{\small $\pm$0.070}} & \hl{\textbf{0.788{\small $\pm$0.004}}} \\
    ResNet flow   & 5.147{\small $\pm$0.171} & \hl{\textbf{6.279{\small $\pm$0.098}}} & \hl{\textbf{0.760{\small $\pm$0.004}}} & \hl{4.903{\small $\pm$0.125}} & \hl{0.784{\small $\pm$0.010}} \\
\end{tabular}

    \caption{Test mean squared error (lower is better) and accuracy/area under curve (higher is better). Best result is bolded, result within one standard deviation is highlighted. Averaged over 5 runs.}
    \label{tab:encoder_decoder_results}
    \vspace{-0.8cm}
\end{table}

\textbf{Filtering approach.}
Following \citet{de2019gru}, we use clinical database \underline{MIMIC-III} \citep{johnson2016mimic}, pre-processed to contain 21250 patients' time series, with 96 features. We also process newly released \underline{MIMIC-IV} \citep{physionet,mimic4} to obtain 17874 patients. The details are in Appendix~\ref{app:mimic}. The goal is to predict the next three measurements in the 12 hour interval after the observation window of 36 hours.

Table~\ref{tab:gru_ode_results} shows that our GRU flow model (\Eqref{eq:gru_continuous}) mostly outperforms GRU-ODE \citep{de2019gru}.
Additionally, we show that the ordinary ResNet flow with $4$ stacked transformations (\Eqref{eq:resnet_flow}) performs worse. The reason might be because it is missing GRU flow properties, such as boundedness. Similarly, an ODE with a regular neural network does not outperform GRU-ODE \citep{de2019gru}. Finally, we report that the model with GRU flow requires 60\% less time to run one training epoch.

\begin{table}
    \centering
    \begin{tabular}{lcccccccccccccc}
                     & \multicolumn{2}{c}{MIMIC-III}         & \multicolumn{2}{c}{MIMIC-IV}                      \\
                     &           MSE     &           NLL     &          MSE    &          NLL                    \\
    \midrule
    GRU-ODE          &  0.507{\small $\pm$0.005}  &\hl{\textbf{0.770{\small $\pm$0.023}}}&\hl{0.379{\small $\pm$0.005}}& 0.748{\small $\pm$0.045}      \\
    ResNet flow      &  0.508{\small $\pm$0.007}  &\hl{0.779{\small $\pm$0.023}}&\hl{0.379{\small $\pm$0.005}}& 0.774{\small $\pm$0.059}                \\
    GRU flow         &\hl{\textbf{0.499{\small $\pm$0.004}}}&\hl{0.781{\small $\pm$0.041}}&\hl{\textbf{0.364{\small $\pm$0.008}}}&\hl{\textbf{0.734{\small $\pm$0.054}}}\\
\end{tabular}

    \caption{Forecasting on healthcare data averaged over 5 runs (lower is better).}
    \label{tab:gru_ode_results}
    \vspace{-0.5cm}
\end{table}

%%% TPP
\textbf{Temporal point processes.}
As we saw in \Secref{sec:tpp}, most of the TPP models consist of two parts: the encoder that processes the history, and the network that outputs the intensity. In the context of neural ODEs, we would like to answer: 1) whether having continuous state $\vh(t)$ outperforms RNNs, and 2) if intertwining the hidden state evolution with the intensity outperforms other approaches. For this purpose we propose the following models based on continuous intensity and mixture distributions.

Jump ODE evolves $\vh(t)$ continuously together with the intensity function $\lambda(t) = g(\vh(t))$ \citep{jia2019jump,chen2021spatio}, where $g$ is a neural network. The neural flow version replaces an ODE with our proposed flow models to evolve $\vh(t)$ and uses Monte Carlo integration to evaluate \Eqref{eq:tpp_likelihood}. Note that this operation can be parallelized unlike solving an ODE.

The mixture-based models keep the same continuous time encoders (ODEs and flows) but output the stationary log-normal mixture for the next arrival time. That is, instead of outputting the continuous intensity, they only use the hidden state at the last observation to define the probability density function \citep{shchur2019intensity}. As a baseline, we use a discrete GRU with the same mixture decoder.

We use both synthetic and real-world data, following \citep{omi2019fully,shchur2019intensity}. We generate 4 synthetic datasets corresponding to homogeneous, renewal and self-correcting processes. For real-world data, we collect timesteps of forum posts (\underline{Reddit}), interactions of students with an online course system (\underline{MOOC}), and \underline{Wiki} page edits  \citep{kumar2019predicting}. The details of the data are in Appendix~\ref{app:tpp}.

We report the test negative log-likelihood on real-world data in Table~\ref{tab:tpp_nll}, for models trained both with and without marks. Full results, including synthetic data can be found in the Appendix~\ref{app:additional_results}. We note that all the models capture the synthetic data, although continuous intensity models struggle compared to those with the mixture distribution. We can see this is the case for real-world data too, where the mixture distribution usually outperforms the corresponding continuous intensity model. In general, neural flows are better than ODE-based models, with the exception of one ODE model on Wiki dataset. We can conclude that having a continuous encoder is preferred to a discrete RNN because it can capture the irregular time sequence better. However, there is no benefit in parametrizing the intensity function in a continuous fashion, especially since this is a much slower approach.

Table~\ref{tab:encoder_decoder_time} in Appendix~\ref{app:additional_results} shows the comparison of wall clock times. Comparing only continuous intensity models we can see that Monte Carlo integration is faster than solving an ODE. As expected, using the mixture distribution gives the best performance. Thus, our flow models offer more than an order of magnitude faster processing compared to ODEs with continuous intensity. \Figref{fig:speed_comparison_all} shows the difference for continuous models on the respective real-world datasets, the gap is even bigger if we include mixture-based models, where the speed-up is over an order of magnitude.

\begin{table}
    \centering
\begin{tabular}{p{0.1cm}lcccccccccc}
                            &&    \multicolumn{2}{c}{MOOC} &   \multicolumn{2}{c}{Reddit} &   \multicolumn{2}{c}{Wiki} \\
    \cmidrule{2-8}
    &Discrete GRU            & -0.4448 & \hl{2.7563}       &  -0.9299 & 1.8468            &  -0.5832 & 8.0527 \\
    \cmidrule{2-8}
    \multirow{3}{*}{\rotatebox{90}{Cont.}}
    &Jump ODE                &  0.8710 & 4.6118            &   0.1308 & 3.6654            & -0.3115 & 10.6040 \\
    &Coupling flow           &  0.7694 & 5.5494            &  -0.1263 & 3.6312            & -0.2807 &  9.7214 \\
    &ResNet flow             &\hl{\textbf{-1.2379}}&2.9466 &\hl{\textbf{-1.2962}}& 2.3932 & -1.2907 & 10.4368 \\
    \cmidrule{2-8}
    \multirow{3}{*}{\rotatebox{90}{Mix.}}
    &Jump ODE                & -0.2626 & 3.0723            & -1.0907 & 1.9057             &\hl{\textbf{-1.3635}} & \hl{\textbf{7.5537}}\\
    &Coupling flow           & -0.4026&\hl{\textbf{2.5877}}& -1.0933 &\hl{\textbf{1.6817}}& -1.2702 & 8.8018 \\
    &ResNet flow             & -0.5664 & 3.0005            & -1.0605 & 1.9491             & -1.1937 & 8.5489 \\
\end{tabular}

    \caption{Test NLL for TPP (left columns, per dataset) and marked TPP (right columns); full results in Appendix~\ref{app:additional_results}. Cont.\ denotes models with continuous intensity, and Mix.\ with mixture distribution.}
    \label{tab:tpp_nll}
    \vspace{-0.5cm}
\end{table}

\textbf{Spatial data.}
We compare the continuous normalizing flows with our continuous-time version of the coupling NF on time-dependent density estimation. We use two versions of each model: time-varying and attentive, as described in \Secref{sec:time_density}.
Following \citet{chen2021spatio}, we use locations of bike rentals (\underline{Bikes}), \underline{Covid} cases for the state of New Jersey \citep{nycovid}, and earthquake events in Japan (\underline{\smash{EQ}}) \citep{geosurvey}.

Results in Table~\ref{tab:stpp_nll} show the test NLL for spatial data, that is, we do not report the TPP loss since this is shared between models. Our continuous coupling NF models perform better on all datasets. Since affine coupling is a simple transformation, we require bigger models with more parameters. At the same time, our models are still more than an order of magnitude faster. Adapting some other, more expressive normalizing flows to satisfy flow constraints might reduce the number of parameters.

\begin{figure}
    \begin{minipage}[b]{0.495\textwidth}
        \centering
        \begin{tabular}{lccc}
                          &Bikes &Covid &EQ \\
    \midrule
    Time-var.\ CNF      &  2.315&  1.984&  1.709\\
    Attentive CNF       &  2.371&  1.973&  1.668\\
    Time-var.\ coupling &\hl{\textbf{2.280}}&\hl{\textbf{1.916}}&  1.633\\
    Attentive coupling  &  2.330&\hl{1.926}&\hl{\textbf{1.457}}\\
\end{tabular}

        \captionof{table}{Test NLL for spatial datasets.}
        \label{tab:stpp_nll}
    \end{minipage}
    \begin{minipage}[b]{0.5\textwidth}
        \centering
        \input{figures/speed_comparison_all.pgf}
        \vspace{-0.6cm}
        \caption{Comparing per-epoch wall-clock times. Each box is dataset (order by appearance in text).}
        % Dataset order: Activity, Physionet, MuJoCo, MIMIC-III, MIMIC-IV, Bikes, Covid, EQ
        \vspace{-0.3cm}
        \label{fig:speed_comparison_all}
    \end{minipage}
\end{figure}

%!TEX root = ../main.tex

\section{Discussion}\label{sec:discussion}

In this paper we presented neural flows as an efficient alternative to neural ODEs. We retain all the desirable properties of neural ODEs, without using numerical solvers. Our method outperforms the ODE based models in time series modeling and density estimation, at a much smaller computation cost. This brings the possibility to scale to larger datasets and models in the future.

\textbf{Other related work.}
Early works on approximating the ODE solutions without numerical solvers used splines or radial basis functions \citep{meade1994numerical,jianyu2002numerical}, or functions similar to modern ResNets \citep{lagaris1998artificial}. More recently, \citep{piscopo2019solving} approximate the solution by minimizing the error of the solution points and of the boundary condition. Unlike these approaches, we do not approximate the solution to some given ODE but learn the solutions which corresponds to learning the unknown ODE. Also, our method guarantees that we always define a proper flow, as is required in certain applications.

A similar problem is modeling the solutions to partial differential equations, e.g., with a model that is analogous to the classical discrete encoder-decoder \citep{li2020fourier}. Although we cannot compare these two settings directly, one could use our method to enhance modeling PDE solutions.

ResNets were initially recognized as a discretization of dynamical systems \citep{liao2016bridging,weinan2017proposal} and were used to tackle infinite depth \citep{bai2019deep,lu2018beyond}, stability \citep{ciccone2018nais,haber2017stable} and invertibility \citep{chang2018reversible,jacobsen2018revnet}. We take a different approach and propose modified ResNets, among other, avoiding any iterative procedure.
ResNets also lead to neural ODEs which have memory efficient backpropagation as one of the main features \citep{farrell2013automated,chen2018neural}. Further, to combat solver inefficiency, many improvements have been proposed, such as adding regularization \citep{finlay2020train,ghosh2020steer,kelly2020learning}, improving training \citep{gholami2019anode,kidger2020hey,zhuang2020adaptive} and having faster inference \citep{poli2020hyper}.

\textbf{Limitations.}
Defining a flow automatically defines an ODE, but since many ODEs do not have closed-form solutions, we cannot always find the \textit{exact} flow corresponding to a particular ODE. This is usually not an issue since in most applications, such as those presented in \Secref{sec:applications}, it is sufficient for both neural ODEs and neural flows to approximate an unknown dynamic.
However, if we restrict ourselves to autonomous ODEs (fixed vector field in time), we cannot define a general neural flow that satisfies this condition. We further discuss this in Appendix~\ref{app:autonomous} and present a potential solution that involves a simple regularization.

Since neural ODEs reuse the same function $f$ in the solver, essentially defining \textit{implicit layers}, they can be more parameter efficient.
Sometimes we might need more parameters to represent the same dynamic, as we observed in the density estimation task. But even here, the results show neural flows are more efficient.
In the special setting with limited memory, we can resort to existing solutions \citep{chen2016training}.

\textbf{Future work.}
In this work we designed neural flow models as invertible functions that satisfy initial condition using simple dependence on time. Although these models already outperform neural ODEs, it would be interesting to see if there are other ways to define a neural flow, and whether these architectures can outperform the ones we proposed here.

We applied our method to the main applications of neural ODEs: time series modeling and density estimation. In the future we hope to see neural flows adapted for other use cases as well. Investigating flows that define the higher order dynamics might also be of interest.

\textbf{Broader impact.}
We introduced a new method to replace neural ODEs. As such, it has a wide variety of potential applications, some of which we explored in this paper. We used several healthcare datasets and hope to see further applications of our method in this domain. At the same time, it is important to pay attention to data privacy and fairness when building such models, especially for sensitive applications, such as healthcare. One of the main benefits of our method is the reduced computation cost, which may imply energy savings.

\newpage

\section*{Acknowledgments}

We would like to thank Oleksandr Shchur for helpful discussions.

\bibliographystyle{abbrvnat}
\bibliography{references.bib}

% \newpage
% \input{sections/checklist}

\newpage
\appendix

%!TEX root = ../main.tex

\section{Theoretical background}

\subsection{GRU-ODE definition}\label{app:gru_ode}

\citet{de2019gru} define the continuous time GRU-ODE model as an ODE that is solved for hidden state $\vh(t)$:
\begin{align}\label{eq:gru_ode}
    \frac{\diff \vh(t)}{\diff t} = (1 - \vz(t)) \odot (\vc(t) - \vh(t)).
\end{align}
With new observation $\vx$, the hidden state is updated with discrete GRU (\Eqref{eq:gru}), and between two observations we solve the ODE given by \Eqref{eq:gru_ode}.

The interesting properties of this model are:
\begin{enumerate}
    \item Boundedness: hidden state $\vh(t)$ stays within range $(-1, 1)$,
    \item Continuity: GRU-ODE is Lipschitz continuous with Lipschitz constant $2$.
\end{enumerate}
In Appendix~\ref{app:gru_flow} we show how our GRU flow model has the same properties without the need to use numerical solvers.

\subsection{Training loss for GRU-ODE-Bayes}\label{app:gru_ode_loss}

\citet{de2019gru} define an objective that mimics the Bayesian filtering. It consists of two parts:
\begin{align}
    \mathcal{L} = \mathcal{L}_{\text{pre}} + \lambda \mathcal{L}_{\text{post}},
\end{align}
where $\mathcal{L}_{\text{pre}}$ is masked negative log-likelihood and $\mathcal{L}_{\text{post}}$ is the Bayesian part of the loss. The model outputs the normal distribution for the observations, conditional on hidden state $\vh(t)$. Since only some features are observed at a time, we mask out the missing values when calculating $\mathcal{L}_{\text{pre}}$. We denote our predicted distribution with $p_\text{pre}$, and predicted distribution after updating the state with $p_\text{post}$. Now, the Bayesian update can be written as $p_{\text{Bayes}} \propto p_{\text{pre}} \cdot p_{\text{obs}}$, with $p_{\text{obs}}$ being the noise of the observations. $\mathcal{L}_{\text{post}}$ is defined as a KL-divergence between $p_{\text{Bayes}}$ and $p_{\text{post}}$. This can be calculated in closed-form for normal distribution.

\subsection{Proof of Theorem~\ref{thm:gru_invertible}}\label{app:gru_flow}

\textbf{Preliminaries.}
Function $f$ has the Lipschitz constant $L$ if $|f(x) - f(y)| \leq L |x - y|$, $\forall x, y$.
We first derive a few useful inequalities.

For the sum of two Lipschitz functions $f + g$, the following holds:
\begin{align}\label{eq:lipschitz_sum}
\begin{split}
    | f(x) + g(x) - f(y) - g(y) |
    &\leq |f(x)-f(y)| + |g(x)-g(y)| \\
    &\leq \Lip(f) |x - y| + \Lip(g) |x - y| \\
    &\leq (\Lip(f) + \Lip(g)) |x - y|,
\end{split}
\end{align}
by the triangle inequality and the definition of the Lipschitz function.
Similarly, for the product of two Lipschitz functions $f \cdot g$, the following holds:
\begin{align}\label{eq:lipschitz_product}
\begin{split}
    | f(x) g(x) - f(y) g(y) |
    &= |f(x)g(x) + f(x)g(y) - f(x)g(y) - f(y) g(y)| \\
    &= | f(x) (g(x) - g(y)) + g(y)(f(x) - f(y)) | \\
    &\leq |f(x)| |g(x) - g(y)| + |g(y)| |f(x) - f(y)| \\
    &\leq |f(x)| \cdot \Lip(g) \cdot |x-y| + |g(y)| \cdot \Lip(f) \cdot |x-y|. \\
    &= (|f(x)| \cdot \Lip(g) + |g(y)| \cdot \Lip(f)) |x-y|.
\end{split}
\end{align}
If $f$ and $g$ are bounded, we can bound the above term too.

Let $f$ be contractive function, $\Lip(f) < 1$. Then, for the composition of functions $\sigmoid \circ f$, where $\sigmoid(x) = (1 + \exp(-x))^{-1}$ is the sigmoid activation, the following holds:
\begin{align*}
    | \sigmoid(f(x)) - \sigmoid(f(y)) | \leq \Lip(\sigmoid) | f(x) - f(y) | = \frac{1}{4} | f(x) - f(y) | \leq \frac{1}{4} |x-y|,
\end{align*}
where we used $\Lip(\sigmoid) = \max(\sigmoid^\prime) = \frac{1}{4}$, by the mean value theorem. Similarly, $\Lip(\tanh) = 1$.

%%% Proof
\begin{proof}(\textbf{Theorem~\ref{thm:gru_invertible}})

\Eqref{eq:gru} defines GRU as: $\vz_t \odot \vh_{t-1} + (1 - \vz_t) \odot \vc_t$. Since $\vz_t$ is defined as $\sigmoid(f_c(\cdot))$, and acts as a gate, we can equivalently define GRU with: $(1 - \vz_t) \odot \vh_{t-1} + \vz_t \odot \vc_t$. This will slightly simplify further calculations. Then, the GRU flow is defined as:
\begin{flalign*}
    && F(t, \vh) = \vh + \Hypernet(t) \odot z(t, \vh) \odot (c(t, \vh) - \vh). && (\text{\ref{eq:gru_continuous}})
\end{flalign*}
$F$ is invertible when the second summand on the right hand side is a contractive map, i.e., has a Lipschitz constant smaller than one. Since $\Hypernet(t)$ is bounded to $[0, 1]$ and does not depend on $\vh$, we only need to show that $z(t, \vh) \odot (c(t, \vh) - \vh)$ is contractive. From here, we denote with $x$ and $y$ the input to our functions.

Following Definition~\ref{def:gru_flow}, let $r(x) = \beta \cdot \sigma(f_r(x))$, with $\Lip(f_r) < 1$. Then we can write:
\begin{align}\label{eq:lip_r}
\begin{split}
    | r(x) - r(y) | &= | \beta \cdot \sigma(f_r(x)) - \beta \cdot \sigma(f_r(y)) | \\
    &\leq \beta |\sigma(f_r(x)) - \sigma(f_r(y))| \\
    &\leq \frac{1}{4}\beta | f_r(x) - f_r(y) | \\
    &< \frac{1}{4}\beta |x-y|.
\end{split}
\end{align}
Similarly, for $z(x)$, where $z(x) = \alpha \cdot \sigma(f_z(x))$, and $\Lip(f_z) < 1$:
\begin{align}\label{eq:lip_z}
\begin{split}
    | z(x) - z(y) | \leq |\alpha \cdot \sigma(f_z(x)) - \alpha \cdot \sigma(f_z(y))| < \frac{1}{4} \alpha |x-y|.
\end{split}
\end{align}
Then for $c(x) = \tanh(f_c(r(x) \cdot x))$, with $\Lip(f_c) < 1$, we can write:
\begin{align}
\begin{split}
    |c(x)-c(y)| &= | \tanh(f_c(r(x) \cdot x)) - \tanh(f_c(r(y) \cdot y))| \\
    &\leq | f_c(r(x) \cdot x) - f_c(r(y) \cdot y) | \\
    &< | r(x) \cdot x - r(y) \cdot y | \\
    &< (\underbrace{|r(x)|}_{< \beta} \cdot \underbrace{\Lip(\mathrm{Id})}_{=1} + \underbrace{|x|}_{< 1} \cdot \underbrace{\Lip(r)}_{< \frac{1}{4}\beta}) |x-y|,
\end{split}
\end{align}
where we used \Eqref{eq:lipschitz_product} in the last line. Then $\Lip(c) < \frac{5}{4}\beta$. Now, for $c(x) - x$, and using \Eqref{eq:lipschitz_sum}, we write:
\begin{align}\label{eq:lip_c}
\begin{split}
    | c(x) - x - c(y) + y | \leq (\Lip(c) + 1) |x - y|,
\end{split}
\end{align}
meaning the whole term has Lipschitz constant $\frac{5}{4}\beta + 1$. Finally, for the term on the right hand side of \Eqref{eq:gru_continuous}, the following holds:
\begin{align*}
    &| z(x)(c(x) - x) - z(y)(c(y) - y) | \\
    &< (
    \underbrace{|z(x)|}_{< \alpha} \cdot
    \underbrace{\Lip(c(x) - x)}_{< \frac{5}{4}\beta + 1}
    +
    \underbrace{|c(x) - x|}_{< 2} \cdot
    \underbrace{\Lip(z(x))}_{< \frac{1}{4} \alpha}
    |x - y| .
\end{align*}
If we set $\alpha = \frac{2}{5}$, $\beta = \frac{4}{5}$, then the Lipschitz constant is smaller than $1$, as required.
\end{proof}
\subsubsection{Properties of GRU flow}
Our GRU flow has the same desired properties as GRU-ODE:
\begin{enumerate}
    \item Boundedness: hidden state $\vh$ stays within range $(-1, 1)$,
    \item Continuity: the whole transformation $\vh + g(\vh)$ has Lipschitz constant $1 + \Lip(g) \leq 2$.
\end{enumerate}
The gating mechanism in discrete GRU helps with gradient propagation to enable learning long-term dependencies. We emphasize that both GRU flow and GRU-ODE update the hidden state in two distinct ways: 1) with discrete GRU when the new observation arrives, and 2) with continuous GRU between observations. Thus, the gates $\vz$ and $\vr$ do not have the same interpretation in discrete GRUCell and in continuous GRU flow or GRU-ODE.

The same way, scalars $\alpha$ and $\beta$ should not be interpreted as bounds to how much information can pass, but as a way to ensure invertibility. GRU flow has the ability to keep the old state $\vh$, and does so at the initial condition $t=0$, but can also overwrite it completely.

\subsection{ODE reparameterization}\label{app:ode_reparameterization}

The ODESolve operation is usually implemented such that it takes scalar start and end times, $t_0$ and $t_1$. However, we are often interested in processing the data in batches, to get speed-up from parallelism on modern hardware. When the previous works \citep{chen2018neural,rubanova2019latent,de2019gru} received the vectors of start and end times, e.g., $\vt_0 = [0, 0, 0]$ and $\vt_1 = [5, 1, 4]$, they would concatenate all the values into a single vector and sort them to get a sequence of strictly ascending times, e.g., $[0, 1, 4, 5]$. The solver would then first solve $0 \rightarrow 1$, then $1 \rightarrow 4$, and finally $4 \rightarrow 5$. Note that for the element in the batch with the largest end time, this requires calling ODESolve multiple times (number of unique time values), instead of only once. Without this procedure, the adaptive solver could take larger steps then the ones imposed by the current batch, meaning we would get better performance.

\citet{chen2021spatio} propose a reparameterization, such that, instead of solving the ODE on the interval $t \in [0, t_{\max}]$, they solve it on $s \in [0, 1]$, with $s = t / t_{\max}$. For the batch of size $n$, the joint system is:
\begin{align*}
    \frac{\diff}{\diff s}
    \begin{bmatrix}
        \vx_1 \\
        \vx_2 \\
        \vdots \\
        \vx_n
    \end{bmatrix} =
    \begin{bmatrix}
        t_1 f(s t_1, \vx_1) \\
        t_2 f(s t_2, \vx_2) \\
        \vdots \\
        t_n f(s t_n, \vx_n)
    \end{bmatrix}.
\end{align*}
This allows solving the system in parallel, in contrast to previous works. We used this reparameterization in all of our experiments.

\subsection{Attentive normalizing flow}\label{app:attentive_nf}

We follow the setup from \Secref{sec:time_density}, denoting times with $\vt=(t_1, \dots, t_n)$, and marks with $\mX = (\vx_1, \dots, \vx_n)$, $\vx_i \in \R^d$. We define the self-attention layer, following \citep{vaswani2017attention}, as:
\begin{align}\label{eq:attention-general}
    \mathrm{SelfAttention}(\mX) = \mathrm{Attention}(\mQ, \mK, \mV) = \softmax \left( \frac{\mQ \mK^T}{\sqrt{d_k}} \right) \mV,
\end{align}
where $\mQ \in \R^{n \times d_k}, \mK \in \R^{n \times d_k}, \mV \in \R^{n \times d_v}$ are matrices that we obtain by transforming each element $\vx_i$ of $\mX$ by a neural network. \citet{chen2021spatio}, in their attentive CNF model, define the function $f$ from \Eqref{eq:cnf} for each $\vx_i$, as the $i$th output of $\mathrm{Attention}$ function. It is important that elements $\vx_j$, $j > i$, do not influence $\vx_i$ to ensure we have a proper temporal model. This is achieved by placing $-\infty$ for values above the diagonal of the $\mQ \mK^T$ matrix so that $\softmax$ returns zero on these places.

Discrete normalizing flows cannot define the transformation using attention and have tractable determinant of the Jacobian at the same time. However, since we actually need an autoregressive model, i.e., the dependence is strictly on the past values, not future, we can define a model similar to attentive CNF. We use \Eqref{eq:attention-general} with diagonal masking to embed the history of all the elements that preceded $\vx_i$: $\vh_i = \mathrm{SelfAttention}(\mX_{1:i-1})$. This is in contrast to \citep{chen2021spatio}, who used $\mX_{1:i}$. Then, the conditioning vector $\vh_i$ is used as an additional input to neural networks $u$ and $v$ from \Eqref{eq:coupling_flow}, essentially defining a conditional affine coupling normalizing flow.

\subsection{Autonomous ODEs}\label{app:autonomous}

Autonomous differential equations are defined with a vector field that is fixed in time $\dot{\vx} = f(\vx(t))$. Note that function $f$ does not depend on time $t$ like before. Therefore, the conditions i) and ii) from \Secref{sec:neural_flows} are not enough to define the corresponding flow. To be precise, the flow $F$ defines an autonomous ODE if it satisfies the additional condition:
\begin{enumerate}
    \setcounter{enumi}{2}
    \item $F(t_1 + t_2, \vx_0) = F(t_2, F(t_1, \vx_0))$,
\end{enumerate}
meaning that solving for $t_1$ first, then $t_2$, is the same as solving for $t_1+t_2$, given initial condition $\vx_0$.

More formally, we defined flow $F$ on set $\R^d$ as a group action of the additive group $G = (\R, +)$ (elements being time points). Equivalently, group action of $G$ on $\R^d$ is a group homeomorphism from $G$ to $\mathrm{Sym}(\R^d)$ (symmetric group, bijective functions and composition $(\Invertible, \circ)$), i.e., some function $\Hypernet : G \rightarrow \mathrm{Sym}(\R^d)$ maps time $t$ to parameters of an invertible neural network $\Invertible$, with $\Hypernet(t_1 + t_2) = \Hypernet(t_1) \circ \Hypernet(t_2)$. Identity element of $G$, $0$ is mapped to an identity function, inverse $-t$ is mapped to an inverse function.

It's clear that our proposed architectures from \Secref{sec:neural_flows} do not satisfy condition iii), unless we redefine it to allow time-dependence. Therefore, one way to satisfy iii) is to have $\frac{\diff}{\diff t}F$ independent of time. Note, however, that if we define the ResNet flow as $\vx_t := F(t, \vx_0) = \vx_0 + t \cdot h(\vx_0)$, then even though time disappears from the derivative $\frac{\diff}{\diff t}F$, the derivative is expressed in terms of $\vx_0$, not $\vx_t$. This means time is still implicitly included since starting at different $\vx_0$ gives different values.

Matrix exponential $\exp(\mA t) \vx$, as a solution to a linear ODE: $\dot{\vx} = \mA \vx$, is one example of a closed-form solution to an autonomous ODE. Another potential \textit{autonomous flow} is of the form $\vx + \Hypernet(t)$, but not $g(\vx) + \Hypernet(t)$, since this does not satisfy initial condition or $g$ must depend on time. To the best of our knowledge, there is no general neural flow parametrization that can capture all autonomous ODEs. Therefore, we can try to learn the desired behavior instead of guaranteeing it.

We can add the penalty to our loss that directly corresponds to condition iii). Given the loss function $\mathcal{L}$ and the current batch of $n$ elements $\mX \in \R^{n\times d}$, $\vt \in \R^{n}$, where we can represent each $t_i \in \vt$ as $t_i = t_i^{(1)} + t_i^{(2)}$, with $t_i^{(1)}, t_i^{(2)}$ uniformly sampled on $[0, t_i]$, the total loss is:
\begin{align}\label{eq:autonomous_reg}
    \mathcal{L}_{\text{total}} = \mathcal{L} + \gamma \frac{1}{n} \sum_i (F(t_i, \vx_i) - F(t_i^{(2)}, F(t_i^{(1)}, \vx_i)))^2 ,
\end{align}
where $\gamma$ is some positive value. The second term penalizes flows that do not satisfy iii), meaning we should get the flow that is closer to the underlying autonomous ODE. This can be calculated in parallel to other computations.

\begin{figure}
    \centering
    \input{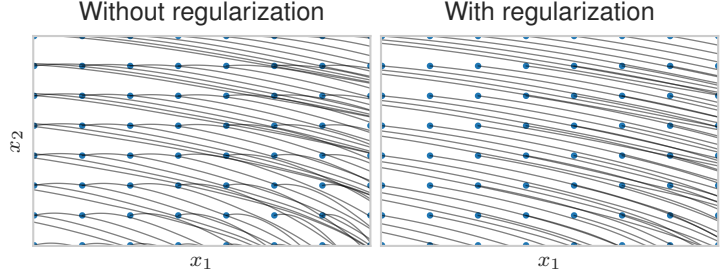}
    \caption{Comparison of the phase space for the same model trained with and without the autonomous regularization (\Eqref{eq:autonomous_reg}). Dots denote initial conditions. Note that the overlapping dynamic does not mean the solutions are not unique, only that the vector field is dependent on time.}
    \label{fig:autonomous_flow}
\end{figure}

\Figref{fig:autonomous_flow} shows the comparison between learning the data generated from an autonomous ODE (see next section for data details), using the regularization as defined in \Eqref{eq:autonomous_reg} and without such regularization. We can see that the base model already learns good behavior but when we include the regularization, the trajectories overlap less frequently.

\subsection{Linear ODE and change of variables}\label{app:linear_ode_density}

Consider a linear ODE $f(t, \vz(t)) = \mA \vz(t)$, with $\vz(0) = \vz$ and $\vz(1)=\vx$. Solving the ODE $0 \rightarrow 1$ is the same as calculating $\exp(\mA) \vz$, where $\exp$ is the matrix exponential. Suppose that $\vz \sim q(\vz)$, then the distribution $p(\vx)$ that we get by transforming $\vx$ with an ODE is defined as:
\begin{align}
    \log p(\vx) = \log q(\vz) - \int_0^1 \mathrm{tr} \left( \frac{\partial f}{\partial \vz(t)} \right) \diff t = \log q(\vz) - \mathrm{tr}(\mA),
\end{align}
or simply: $p(\vx) = q(\vz) \exp(\mathrm{tr}(\mA))^{-1}$.

When using the Hutchinson's trace estimator for the trace approximation we get the same result: $\E_{p(\vepsilon)}[\int_0^1 \vepsilon^T \frac{\partial f}{\partial \vz(t)} \vepsilon \diff t] = \E_{p(\vepsilon)} [\vepsilon^T \mA \vepsilon] = \mathrm{tr}(\mA)$, where $\E(\vepsilon) = \vzero$ and $\mathrm{Cov}(\vepsilon) = \mI$.

Similarly, applying the discrete change of variables, we get the same result for the matrix exponential:
\begin{align}
    p(\vx) = q(\vz) | \det J_F(\vz) |^{-1} = q(\vz) | \det \exp(\mA) |^{-1} = q(\vz) \exp(\mathrm{tr}(\mA))^{-1}.
\end{align}

\subsection{Computation complexity of (continuous) normalizing flows}\label{app:trace_complexity}

In general, evaluating the trace of the Jacobian of function $f:\R^d \rightarrow \R^d$ requires $O(d^2)$ operations. In CNFs, this operation has to be performed at every solver step. Since the number of steps can be very large for more complicated distributions \citep{grathwohl2018ffjord}, this becomes prohibitively expensive. Because of this, \citet{grathwohl2018ffjord} introduce computing the approximation of the trace during training. This has the benefit of having a lower cost, $O(d)$. The issue with this method is that the training becomes noisier and after training we have to again rely on exact trace to get the exact density.

On the other hand, computing the determinant of the Jacobian is $O(d^3)$ operation in general. Because of this, regular normalizing flows do not use unconstrained functions $f$, but rather opt for those that produce triangular Jacobians, e.g., autoregressive \citep{kingma2016improved} or coupling transformations \citep{dinh2016density}, where the determinant is just the product of the diagonal elements, i.e., it is of linear cost $O(d)$.

%!TEX root = ../main.tex

\begin{figure}
    \centering
    \input{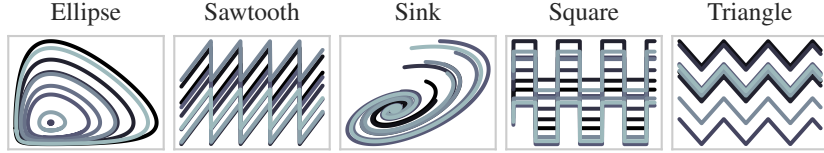}
    \caption{Sample trajectories for synthetic data.}
    \label{fig:synthetic_data}
\end{figure}

\section{Synthetic experiments}\label{app:synthetic_experiment}

We first test the capabilities of our models on periodic signals:
\begin{itemize}
    \item Sine: $f(t, x) = \cos(t)$ which corresponds to flow $F(t, x) = x + \sin(t)$, $x \in \R$,
    \item Sawtooth: $F(t, x) = x + t - \lfloor t \rfloor$,
    \item Square: $F(t, x) = x + \sign(\sin(t))$,
    \item Triangle: $F(t, x) = \int_0^t \sign(\sin(u))\diff u$.
\end{itemize}
We sample initial values $x$ uniformly in $(-2, 2)$ and set the time interval to $(0, 10)$. We additionally check how well the models extrapolate by extending the initial condition interval to $(-4, 4)$ and time to $30$. We also use two datasets, generated as solutions to known ODEs:
\begin{itemize}
    \item Sink: $f(t, \vx) =
    \begin{bmatrix}
        -4 & 10 \\
        -3 & 2
    \end{bmatrix}
    \begin{bmatrix}
        x_1 \\ x_2
    \end{bmatrix}$,
    \item Ellipse: $f(t, \vx) =
    \begin{bmatrix}
        \frac{2}{3} x_1 - \frac{2}{3} x_1  x_2\\
        x_1 x_2 - x_2
    \end{bmatrix}$, which is a particular parametrization of Lotka-Volterra equations, also known as predator-prey equations,
\end{itemize}
where we sample initial conditions $x_1, x_2 \in [0, 1]$ uniformly. For extrapolation, we use $x_1, x_2 \in [1, 2]$.
\Figref{fig:synthetic_data} shows the generated trajectories for all synthetic datasets.

\begin{figure}
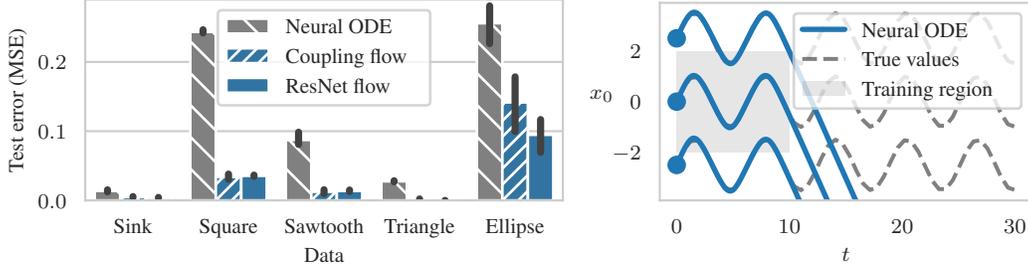

    \centering
    \input{figures/synthetic_results.pgf}
    \input{figures/sine_extrapolation.pgf}
    \caption{(Left) Test error for synthetic data. (Right) All models fail when extrapolating in time.}
    \label{fig:synthetic_results}
\end{figure}

\subsection{Comparing adaptive and fixed-step solvers}\label{app:solver_comparison}

\begin{figure}
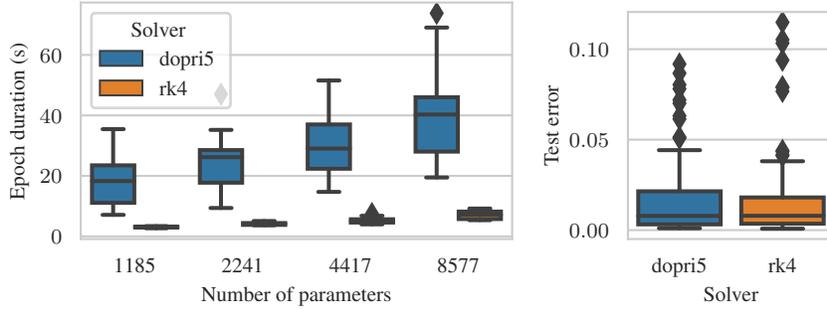

    \centering
    \input{figures/ode_sine_epoch_duration.pgf}
    \input{figures/ode_sine_loss.pgf}
    \caption{Fixed solvers are faster to train on synthetic data (Left) but they still have similar accuracy compared to adaptive solvers (Right).}
    \label{fig:ode_sine_epoch_duration}
\end{figure}

We ran an extensive hyperparameter search for Sine dataset. We test models with $2$ or $3$ hidden layers, each having dimension of $32$ or $64$, use $\tanh$ or $\mathrm{ELU}$ activations between them, and have $\tanh$ or identity as the final activation. For each of the model configurations we apply either no regularization or weigh the penalty term with $10^{-3}$. Finally, we run each trial $5$ times with different seeds and compare between Runge-Kutta fixed-step solver with $20$ steps and an adaptive 5th order Dormand-Prince method \citep{dormand1980family}.

As expected, the vast majority of the trials fit the data very well. However, as \Figref{fig:ode_sine_epoch_duration} shows, an adaptive solver always requires significantly longer training times, regardless of the size of the model, choice of the activations or regularization. We used default tolerance settings ($\mathrm{rtol} = 10^{-7}$, $\mathrm{atol} = 10^{-9}$) which is why we get such long training times. Therefore, in the other experiments, in the main text, whenever we use dopri5, we use $\mathrm{rtol} = 10^{-3}$ and $\mathrm{atol} = 10^{-4}$ to make training feasible. This once again shows the trade-off between speed and numerical accuracy.

\begin{wrapfigure}[12]{r}{0.43\textwidth}
    \centering
    \includegraphics{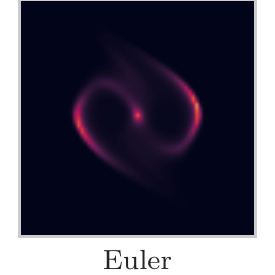}
    \includegraphics{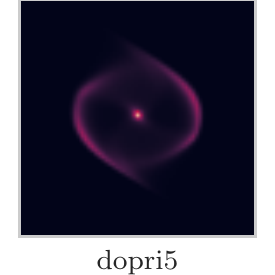}
    \caption{Density learned with Euler and dopri5 solver. The estimated area under the curve for Euler method is $1.06$, meaning it does not define a proper density.}
    \label{fig:nf_solver_comparison}
\end{wrapfigure}

From the results, one would expect that we can safely use fixed-step solvers and achieve similar or better results with smaller computational demand. However, as \citet{ott2020neural} showed, this can lead to overlapping trajectories which give non-unique solutions. Breaking the assumptions of our model can lead to misleading results in some cases. Here, we tackle density estimation with continuous normalizing flows as an example.

We construct a synthetic 2-dim.\ dataset as a mixture of zero-centered normal distribution ($\sigma = 0.05$) and uniform points on the perimeter of a unit circle with small noise ($\sigma = 0.01$). We test adaptive dopri5 solver and Euler method with 20 steps.

The fixed solver achieves better results but \Figref{fig:nf_solver_comparison} visually demonstrates that it is not really capturing the true distribution better. It \textit{cheats} by not defining a proper density function that integrates to $1$. Since it has more mass to distribute, it can achieve better results. This might be hard to detect in higher dimensions and it can be particularly problematic since most of the literature reports log-likelihood on test data. Even though we took Euler method as an extreme example, the same can be shown for other solvers as well.

\subsection{Comparing flow configurations}

Similar to Appendix~\ref{app:solver_comparison}, we compare different flow models on synthetic sine data. We try coupling and ResNet models with linear and $\tanh$ for $\Hypernet$, as well as an embedding with $8$ Fourier features (bounded to $(0, 1)$ interval in ResNet model), see \Secref{sec:proposed_architectures} for more details. Both models have either $2$ or $4$ stacked transformations, each with a two hidden layer neural network with $64$ hidden dimensions. We run each configuration $5$ times with and without weight regularization ($10^{-3}$).

All the models capture the data perfectly, except for the coupling flow with linear function of time $\Hypernet$ which does not converge. This could be due to inability of neural networks to process large input values. The issue can be fixed with different initialization or normalizing the input time values.

Tables~\ref{tab:synthetic_results}, \ref{tab:synthetic_space_results} and \ref{tab:synthetic_duration_results} show that neural flows outperform neural ODEs in forecasting, extrapolation with different initial values, and are faster during training.

\begin{table}[t]
    \centering
    \begin{tabular}{lccccc}
    MSE ($\times 10^{-2}$)& Ellipse &           Sawtooth &               Sink &             Square &           Triangle \\
    \midrule
    Neural ODE    &  25.59{\scriptsize $\pm$3.19} &  8.74{\scriptsize $\pm$1.10} &  1.38{\scriptsize $\pm$0.17} &   24.34{\scriptsize $\pm$0.3} &  2.76{\scriptsize $\pm$0.09} \\
    Coupling flow &   14.16{\scriptsize $\pm$4.80} &\textbf{1.25{\scriptsize $\pm$0.33}}& 0.50{\scriptsize $\pm$0.06} &\textbf{3.38{\scriptsize $\pm$0.4}}&  0.19{\scriptsize $\pm$0.02} \\
    ResNet flow   &\textbf{9.48{\scriptsize $\pm$2.64}}&  1.38{\scriptsize $\pm$0.13} &\textbf{0.40{\scriptsize $\pm$0.04}}&   3.56{\scriptsize $\pm$0.1} &\textbf{0.0{\scriptsize $\pm$0.0}} \\
\end{tabular}

    \caption{Test error on synthetic data, lower is better. Best results in bold.}
    \label{tab:synthetic_results}
\end{table}

\begin{table}[t]
    \centering
    \begin{tabular}{lccccccc}
    MSE ($\times 10^{-2}$)&    Ellipse &        Sawtooth &           Sink &         Square &       Triangle \\
    \midrule
    Neural ODE    & \textbf{19.82{\scriptsize {\scriptsize $\pm$1.34}}}&  10.64{\scriptsize {\scriptsize $\pm$1.76}} &  18.0{\scriptsize {\scriptsize $\pm$1.18}} &  32.96{\scriptsize {\scriptsize $\pm$3.0}} &  4.22{\scriptsize {\scriptsize $\pm$0.56}} \\
    Coupling flow &  515.8{\scriptsize $\pm$555.6} &\textbf{1.32{\scriptsize {\scriptsize $\pm$0.36}}} &\textbf{5.53{\scriptsize {\scriptsize $\pm$2.23}}} &\textbf{3.93{\scriptsize {\scriptsize $\pm$0.76}}} &\textbf{0.2{\scriptsize {\scriptsize $\pm$0.04}}} \\
    ResNet flow   &   100.4{\scriptsize $\pm$45.4} &   3.49{\scriptsize {\scriptsize $\pm$1.14}} &  6.65{\scriptsize {\scriptsize $\pm$2.23}} &  9.84{\scriptsize {\scriptsize $\pm$2.94}} &  0.79{\scriptsize {\scriptsize $\pm$0.21}} \\
\end{tabular}

    \caption{Error on trajectories that start at initial conditions out of training distribution. Some trials returned outliers that skew the results (e.g., coupling flow on ellipse dataset).}
    \label{tab:synthetic_space_results}
\end{table}

\begin{table}[t!]
    \centering
    \begin{tabular}{lccccccc}
                  &        Ellipse &       Sawtooth &           Sink &         Square &       Triangle \\
    \midrule
    Neural ODE    &   9.3{\scriptsize $\pm$0.88} &  8.25{\scriptsize $\pm$0.33} &  8.78{\scriptsize $\pm$0.81} &  7.81{\scriptsize $\pm$0.34} &  7.91{\scriptsize $\pm$0.35} \\
    Coupling flow &\textbf{0.7{\scriptsize $\pm$0.11}}&\textbf{0.46{\scriptsize $\pm$0.22}} &\textbf{0.6{\scriptsize $\pm$0.05}} &\textbf{0.49{\scriptsize $\pm$0.14}} &\textbf{0.58{\scriptsize $\pm$0.16}} \\
    ResNet flow   &  1.05{\scriptsize $\pm$0.04} &  1.01{\scriptsize $\pm$0.15} &  1.24{\scriptsize $\pm$0.13} &  0.98{\scriptsize $\pm$0.04} &  1.01{\scriptsize $\pm$0.09} \\
\end{tabular}

    \caption{Wall-clock time (in seconds) to run the last training epoch, using the same batch size.}
    \label{tab:synthetic_duration_results}
\end{table}

\section{Additional results}\label{app:additional_results}

Table~\ref{tab:encoder_decoder_time} compares the training times for smoothing experiment. Neural ODE models use Euler method with 20 steps (the adaptive method is slower).
Table~\ref{tab:tpp_time} shows the average wall-clock time to run a single epoch for different TPP models. We include ablations for flow and ODE models that use different continuous RNN encoders, and a model without an encoder.
Table~\ref{tab:tpp_nll_full} shows full negative log-likelihood results for the TPP experiment.
Table~\ref{tab:tpp_marks_nll} shows the full NLL results for marked TPPs.

\begin{table}[h!]
    \centering
    \begin{tabular}{lcccccccc}
                  &           Activity &              MuJoCo &          Physionet \\
    \midrule
    Neural ODE    &  200.884{\scriptsize $\pm$7.239} &  192.209{\scriptsize $\pm$2.526} &  103.198{\scriptsize $\pm$4.977} \\
    Coupling flow &\textbf{106.298{\scriptsize $\pm$2.314}}&\textbf{46.171{\scriptsize $\pm$1.742}}&\textbf{78.561{\scriptsize $\pm$1.050}}\\
    ResNet flow   &  134.336{\scriptsize $\pm$3.453} &  102.745{\scriptsize $\pm$2.369} &  101.966{\scriptsize $\pm$8.285} \\
\end{tabular}

    \caption{Average time (in seconds) to run a single epoch during training for different models, all other training parameters being the same.}
    \label{tab:encoder_decoder_time}
\end{table}

\begin{table}[h!]
    \centering
    \begin{tabular}{llccccccc}
    && Poisson& Hawkes1 &Hawkes2 & Renewal &  MOOC &Reddit &    Wiki \\
    \cmidrule{2-9}
    \multirow{3}{*}{\rotatebox{90}{Cont.}}
    & Neural ODE              &   96.7 &   129.8 &  208.6 &   111.2 & 844.2 & 612.8 & 157.9 \\
    & Coupling flow           &   10.8 &    11.2 &   10.8 &    11.1 & 180.8 & 113.1 &  31.7 \\
    & ResNet flow             &    7.1 &     7.1 &    7.2 &     7.3 & 130.0 &  83.8 &  19.9 \\
    \cmidrule{2-9}
    \multirow{5}{*}{\rotatebox{90}{Mixture}}
    & GRU-ODE                 &   39.7 &    42.3 &   55.9 &    39.3 & 600.0 & 419.5 &  97.9 \\
    & ODE-LSTM                &   35.9 &    39.0 &   37.8 &    43.8 & 569.4 & 443.6 & 109.4 \\
    & Coupling flow           &    3.4 &     3.4 &    3.3 &     3.3 &  47.0 &  37.2 &   8.5 \\
    & ResNet flow             &    5.9 &     5.9 &    5.8 &     5.9 &  96.5 &  64.9 &  16.1 \\
    & GRU flow                &    3.6 &     3.5 &    3.3 &     3.7 &  52.8 &  36.4 &   9.7 \\
\end{tabular}

    \caption{Average time (in seconds) to run a single epoch during training for TPP models.}
    \label{tab:tpp_time}
\end{table}

\begin{table}[h]
    \centering
    \begin{tabular}{p{0.1cm}lccccccc}
    \multicolumn{2}{l}{Synthetic data} &Poisson &           Hawkes1 &           Hawkes2 &          Renewal \\
    \cmidrule{2-6}
    &Ground truth            &  0.9996 &  0.6405 &  0.1192 &  0.2667 \\
    &Without history& \hl{1.0046} &  0.7826 &  0.2354 &  0.2837 \\
    &Discrete GRU   &\hl{1.0097{\scriptsize $\pm$0.005}}&\hl{0.6424{\scriptsize $\pm$0.006}}&\hl{0.1267{\scriptsize $\pm$0.006}}& \hl{\textbf{0.2598{\scriptsize $\pm$0.016}}} \\
    \cmidrule{2-6}
    \multirow{3}{*}{\rotatebox{90}{Cont.}}
    &Jump ODE       &\hl{\textbf{0.9945{\scriptsize {\scriptsize $\pm$0.016}}}}&\hl{0.6461{\scriptsize $\pm$0.009}}&  0.2246{\scriptsize $\pm$0.042} & 0.3124{\scriptsize $\pm$0.022} \\
    &Coupling flow  &\hl{1.0099{\scriptsize $\pm$0.005}}&\hl{0.6441{\scriptsize $\pm$0.007}}&  0.1376{\scriptsize $\pm$0.005} & \hl{0.2720{\scriptsize $\pm$0.017}} \\
    &ResNet flow    &\hl{1.0105{\scriptsize $\pm$0.005}}&\hl{0.6426{\scriptsize $\pm$0.007}}&  0.1813{\scriptsize $\pm$0.025} & 0.2851{\scriptsize $\pm$0.018} \\
    \cmidrule{2-6}
    \multirow{4}{*}{\rotatebox{90}{Mix.}}
    &GRU-ODE        &\hl{1.0100{\scriptsize $\pm$0.005}}&\hl{\textbf{0.6419{\scriptsize $\pm$0.007}}}&\hl{\textbf{0.1239{\scriptsize $\pm$0.005}}} & \hl{0.2601{\scriptsize $\pm$0.017}} \\
    &ODE-LSTM       &  1.0108{\scriptsize $\pm$0.005} &\hl{0.6448{\scriptsize $\pm$0.006}}&\hl{0.1253{\scriptsize $\pm$0.005}} & \hl{0.2605{\scriptsize $\pm$0.017}} \\
    &Coupling flow  &\hl{1.0103{\scriptsize $\pm$0.005}} &\hl{0.6450{\scriptsize $\pm$0.008}}&\hl{0.1254{\scriptsize $\pm$0.006}} & \hl{0.2605{\scriptsize $\pm$0.016}} \\
    &GRU flow       &\hl{1.0100{\scriptsize $\pm$0.005}} &\hl{0.6439{\scriptsize $\pm$0.007}}&\hl{0.1270{\scriptsize $\pm$0.006}} & \hl{0.2608{\scriptsize $\pm$0.016}} \\
    &ResNet flow    &\hl{1.0104{\scriptsize $\pm$0.005}} &\hl{0.6443{\scriptsize $\pm$0.006}}&\hl{0.1249{\scriptsize $\pm$0.005}} & \hl{0.2603{\scriptsize $\pm$0.017}} \\
    \\
    \multicolumn{2}{l}{Real-word data}&MOOC &             Reddit &              Wiki \\
    \cmidrule{2-5}
    &Without history&   2.0623 &   1.5402 & 1.5813  \\
    &Discrete GRU   &  -0.4448{\scriptsize $\pm$0.294} &  -0.9299{\scriptsize $\pm$0.118} & -0.5832{\scriptsize $\pm$0.321} \\
    \cmidrule{2-5}
    \multirow{3}{*}{\rotatebox{90}{Cont.}}
    &Jump ODE       &   0.8710{\scriptsize $\pm$0.157} &   0.1308{\scriptsize $\pm$0.018} & -0.3115{\scriptsize $\pm$0.011} \\
    &Coupling flow  &   0.7694{\scriptsize $\pm$0.172} &  -0.1263{\scriptsize $\pm$0.273} & -0.2807{\scriptsize $\pm$0.500} \\
    &ResNet flow    &\hl{\textbf{-1.2379{\scriptsize $\pm$0.049}}}&\hl{\textbf{-1.2962{\scriptsize $\pm$0.126}}}&\hl{-1.2907{\scriptsize $\pm$0.045}}\\
    \cmidrule{2-5}
    \multirow{4}{*}{\rotatebox{90}{Mix.}}
    &GRU-ODE        &  -0.2626{\scriptsize $\pm$0.183} &  -1.0907{\scriptsize $\pm$0.076} &\hl{-1.3635{\scriptsize $\pm$0.071}}\\
    &ODE-LSTM       &  -0.2277{\scriptsize $\pm$0.331} &  -1.0888{\scriptsize $\pm$0.029} &\hl{\textbf{-1.3727{\scriptsize $\pm$0.327}}}\\
    &Coupling flow  &  -0.4026{\scriptsize $\pm$0.584} &  -1.0933{\scriptsize $\pm$0.161} &\hl{-1.2702{\scriptsize $\pm$0.178}}\\
    &GRU flow       &  -0.3509{\scriptsize $\pm$0.220} &  -1.0605{\scriptsize $\pm$0.113} & -0.9852{\scriptsize $\pm$0.105} \\
    &ResNet flow    &  -0.5664{\scriptsize $\pm$0.278} &  -1.0291{\scriptsize $\pm$0.174} &\hl{-1.1937{\scriptsize $\pm$0.048}}\\
\end{tabular}

    \caption{Test negative log-likelihood (mean$\pm$standard deviation) for all TPP models.}
    \label{tab:tpp_nll_full}
\end{table}

\begin{table}[h]
    \centering
    \begin{tabular}{p{0.1cm}lccc}
    &&              MOOC &            Reddit &                    Wiki \\
    \cmidrule{2-5}
    &Discrete GRU   &\hl{2.7563{\scriptsize $\pm$0.141}}&  1.8468{\scriptsize $\pm$0.016} &         8.0527{\scriptsize $\pm$0.170}\\
    \cmidrule{2-5}
    \multirow{3}{*}{\rotatebox{90}{Cont.}}
    &Jump ODE       &  4.6118{\scriptsize $\pm$0.070} &  3.6654{\scriptsize $\pm$0.000} &       10.6040{\scriptsize $\pm$0.304} \\
    &Coupling flow  &  5.5494{\scriptsize $\pm$0.413} &  3.6312{\scriptsize $\pm$0.324} &        9.7214{\scriptsize $\pm$0.101} \\
    &ResNet flow    &  2.9466{\scriptsize $\pm$0.000} &  2.3932{\scriptsize $\pm$0.131} &       10.4368{\scriptsize $\pm$0.034} \\
    \cmidrule{2-5}
    \multirow{4}{*}{\rotatebox{90}{Mix.}}
    &GRU-ODE        &  3.5344{\scriptsize $\pm$0.242} &  2.3078{\scriptsize $\pm$0.033} &\hl{\textbf{7.5537{\scriptsize $\pm$0.065}}}\\
    &ODE-LSTM       &  3.0723{\scriptsize $\pm$0.114} &  1.9057{\scriptsize $\pm$0.164} &        8.3187{\scriptsize $\pm$0.231} \\
    &Coupling flow  &\hl{\textbf{2.5877{\scriptsize $\pm$0.176}}}&\hl{\textbf{1.6817{\scriptsize $\pm$0.095}}}&        8.8018{\scriptsize $\pm$0.057} \\
    &ResNet flow    &  3.0005{\scriptsize $\pm$0.081} &  1.9491{\scriptsize $\pm$0.008} &        8.5489{\scriptsize $\pm$0.267} \\
\end{tabular}

    \caption{Test negative log-likelihood (mean$\pm$standard deviation) for all marked TPP models.}
    \label{tab:tpp_marks_nll}
\end{table}

\section{Data pre-processing}

\subsection{Encoder-decoder datasets}\label{app:encoder_decoder}

\textbf{MuJoCo dataset.} Using Deep Mind Control Suite and MuJoCo simulator, \citet{rubanova2019latent} generate 10000 sequences by sampling initial body position in $\R^2$ uniformly from $[0, 0.5]$, limbs from $[-2, 2]$, and velocities from $[-5, 5]$ interval. We use this dataset without any changes.

\textbf{Activity dataset.}
Following \citep{rubanova2019latent}, we round up the time measurements to 100ms intervals. This was done to reduce the size of the union of all the points when batching but is unnecessary when using our flow models, and also when using the reparameterization for ODEs \citep{chen2021spatio}.

Original labels are: walking, falling, lying down, lying, sitting down, sitting, standing up from lying, on all fours, sitting on the ground, standing up from sitting, standing up from sitting on the ground. \citet{rubanova2019latent} combine similar positions into one group resulting in 7 classes: walking, falling, lying, sitting, standing up, on all fours, sitting on the ground. Data is split in train, validation and test set (75\%--5\%--20\%).

\textbf{Physionet dataset.} We use PhysioNet Challenge 2012 \citep{silva2012predicting}, where the goal is to predict the mortality of patients upon being admitted to ICU. We process the data following \citep{rubanova2019latent} to exclude time-invariant features, and round the time stamps to one minute. Each feature is normalized to $[0, 1]$ interval. Data is split the same way as for MuJoCo: 60\%--20\%--20\%.

When reporting MSE scores for the reconstruction task we scale the result by $10^2$ for activity dataset and by $10^3$ for others, for better readability. This is equivalent to scaling the data beforehand.

\subsection{MIMIC-III and MIMIC-IV}\label{app:mimic}

We follow \citep{de2019gru} for processing \textbf{MIMIC-III} dataset. We process MIMIC-IV in a similar vein.

The publicly available \textbf{MIMIC-IV} database provides clinical data of intensive care unit (ICU) patients at the tertiary academic medical center in Boston \citep{mimic4,physionet}. It builds upon the MIMIC-III database and contains de-identified patient records from 2008 to 2019 \citep{johnson2016mimic}. We use version MIMIC-IV 1.0, which was released March 16th, 2021.

To preprocess the data, we first select the subset of patients who:
\begin{itemize}
\item are registered in the admissions table,
\item stayed in the ICU for at least 2 days and no more than 30 days,
\item are older than 15 years at the time of the admission, and
\item have chart-event data available,
\end{itemize}
which leaves us with 17874 patients.

There are four types of data sources available for ICU patients: chart-events, inputs, outputs and prescriptions. The chart-events table contains the patient's routine vital signs as well as any additional information such as laboratory tests. The input table documents drugs administered to the patient through, e.g., solutions and the prescription table stores information about medication given in any other form. Lastly, the outputs table contains any output data from, e.g., a catheter for the patient during their ICU stay.

Because the medication in the input table is administered over time, the administered units and doses have to be unified and then split into entries which are spread out over time. We choose 30 minutes as our sampling window and, for all administered medications with duration longer than an hour, split them into fixed time injections.

For all other tables, we only keep the most commonly used entries:
\begin{itemize}
    \item \textbf{Chart-events}: Alanine Aminotransferase, Albumin, Alkaline Phosphatase, Anion Gap, Asparate Aminotransferase, Base Excess, Basophils, Bicarbonate, Bilirubin, Calcium, Chloride, Creatinine, Eosinophils, Glucose, Hematocrit, Hemoglobin, Lactate, Lymphocytes, Magnesium, MCH, MCV, Monocytes, Neutrophils, pCO2, pH, pO2, Phosphate, Platelet Count, Potassium, PT, PTT, RDW, Red Blood Cells, Sodium, Specific Gravity, Total CO2, Urea Nitrogen and White Blood Cells.
    \item \textbf{Outputs}: Chest Tube, Emesis, Fecal Bag, Foley, Jackson Pratt, Nasogastric, OR EBL, OR Urine, Oral Gastric, Pre-Admission, Stool, Straight Cath, TF Residual, TF Residual Output and Void.
    \item \textbf{Prescriptions}: Acetaminophen, Aspirin, Bisacodyl, D5W, Docusate Sodium, Heparin, Humulin-R Insulin, Insulin, Magnesium Sulfate, Metoprolol Tartrate, Pantoprazole, Potassium Chloride and Sodium Chloride 0.9\% Flush.
\end{itemize}

\subsection{TPP datasets}\label{app:tpp}

We follow previous works to generate and pre-process temporal point process data \citep{omi2019fully,shchur2019intensity,kumar2019predicting}.

\textbf{Synthetic data.} We use 4 synthetic datasets, for each we generate 1000 sequences, each sequence containing 100 elements. We generate Poisson dataset with constant intensity $\lambda^*(t) = 1$; Renewal with stationary log-normal density function ($\mu=1$, $\sigma=6$); and two Hawkes datasets with the conditional intensity $\lambda^*(t) = \mu + \sum_{t_i<t}\sum_j^M \alpha_j\beta_j\exp(-\beta_j(t - t_i))$, with $M=1$, $\mu = 0.02$, $\alpha = 0.8$ and $\beta = 1$ (Hawkes1), or $M = 2$, $\mu = 0.2$, $\alpha = [0.4, 0.4]$ and $\beta = [1, 20]$ (Hawkes2).

\textbf{Reddit.} We use timestamps of posts from most active users to most active topic boards (subreddits) \citep{kumar2019predicting}. There are 984 unique subreddits that we use as marks. We have 1000 sequences in total, each sequence is truncated to contain at most 100 points. This is done to make training with ODE-based models feasible.

\textbf{MOOC} is a dataset containing timestamps of events performed by users in interaction with a learning platform \citep{kumar2019predicting}. There are 7047 sequences, with at most 200 events. We have 97 different mark types corresponding to different interaction types.

\textbf{Wiki} contains timestamps of edits of most edited pages from most active users \citep{kumar2019predicting}. There are 1000 pages (sequences) with at most 250 events, and 984 users that we use as marks.

In our implementation, we use inter-event times $\tau_i = t_i - t_{i-1}$ and for real-world data, we normalize them by dividing them with the empirical mean $\bar{\tau}$ from the training set $\tau_i \mapsto \tau_i / \bar{\tau}$. This can still yield quite large values so for better numerical stability during training, we use log-transform $\tau \mapsto \log(\tau + 1)$. We can think of log-transformation as a change of variables and include it in the negative log-likelihood loss using the probability change of variables formula (see \Secref{sec:time_density}).

\subsection{Spatial datasets}

For spatial data used in time-dependent density estimation experiment, we used the datasets from \citet{chen2021spatio} with the same pre-processing pipeline. See \citep{chen2021spatio} for further details.

\textbf{Earthquakes} contains earthquakes gathered between 1990 and 2020 in Japan, with the magnitude of at least 2.5 \citep{geosurvey}. Each sequence has length of 30 days, with the gap of 7 days between sequences. There are 950 training sequences, and 50 validation and test sequences.

\textbf{Covid} data uses daily cases from March to July 2020 in New Jersey state \citep{nycovid}. The data is gathered on county level and dequantized. Each sequence covers 7 days. There are 1450 sequences in the training set, 100 in validation and 100 in test set.

\textbf{Bikes} contains rental events from a bike sharing service in New York using data from April to August 2019. Each sequence corresponds to a single day, starting at 5am. The data is split in training, test and validation set: 2440, 300, 320 sequences, respectively.

All the spatial values are normalized to zero mean and unit variance. We also normalize the temporal component to $[0, 1]$ interval.

\section{Hyperparameters}

All experiments: Adam optimizer, with weight decay 1e-4

\textbf{Smoothing experiments}
\begin{itemize}
    \item Batch size: 100
    \item Learning rate: 1e-3 with decay 0.5 every 20 epochs
    \item Hidden layers: 3
    \item Models
    \begin{itemize}
        \item \underline{\smash{ODE models}}
        \begin{itemize}
            \item Solver: euler or dopri5
        \end{itemize}
        \item \underline{\smash{Flow models}}: ResNet or coupling flow
        \begin{itemize}
            \item Flow layers: 1 or 2
            \item $\Hypernet(t)$: tanh for ResNet and linear for coupling (used in all experiments)
        \end{itemize}
    \end{itemize}
    \item Datasets
    \begin{itemize}
        \item MuJoCo
        \begin{itemize}
            \item Encoder-decoder hidden dimension: 100-100
            \item Latent state dimension: 20
            \item GRU dimension: 50
        \end{itemize}
        \item Activity
        \begin{itemize}
            \item Encoder-decoder hidden dimension: 30-100
            \item Latent state dimension: 20
            \item GRU dimension: 100
        \end{itemize}
        \item Physionet
        \begin{itemize}
            \item Encoder-decoder hidden dimension: 40-50
            \item Latent state dimension: 20
            \item GRU dimension: 50
        \end{itemize}
    \end{itemize}
\end{itemize}

\textbf{Filtering experiment}
\begin{itemize}
    \item Batch size: 100
    \item Learning rate: 1e-3 with decay 0.33 every 20 epochs
    \item Hidden dimension: 64
    \item Datasets: MIMIC-III or MIMIC-IV
    \item \underline{\smash{ODE models}}
    \begin{itemize}
        \item Solver: euler or dopri5
        \item Hidden layers: 3
    \end{itemize}
    \item \underline{\smash{Flow models}}: GRU flow or ResNet flow
    \begin{itemize}
        \item Flow layers: 1 or 4
        \item Hidden layers: 2
    \end{itemize}
\end{itemize}

\textbf{TPP experiment} (With or without marks)
\begin{itemize}
    \item Batch size: 50
    \item Learning rate: 1e-3
    \item Hidden dimension: 64
    \item Data: Reddit or MOOC or Wiki
    \item \underline{\smash{ODE models}}
    \begin{itemize}
        \item Models: continuous or mixture
        \begin{itemize}
            \item Mixture models: ODE-LSTM or GRU-ODE
        \end{itemize}
        \item Hidden layers: 3
    \end{itemize}
    \item \underline{\smash{Flow models}}
    \begin{itemize}
        \item Models: continuous or mixture
        \begin{itemize}
            \item Continuous models: ResNet or coupling flow
            \item Mixture models: ResNet or coupling or GRU flow
        \end{itemize}
        \item Flow layers: 1
        \item Hidden layers: 2
    \end{itemize}
    \item \underline{\smash{RNN models}}: GRU

\end{itemize}

\textbf{Density estimation experiment}
\begin{itemize}
    \item Batch size: 50
    \item Learning rate: 1e-3
    \item Hidden dimension: 64
    \item Models: time-varying or attentive (for both CNFs and NFs)
    \item \underline{\smash{Continuous normalizing flows}}
    \begin{itemize}
        \item Hidden layers: 4
    \end{itemize}
    \item \underline{\smash{Coupling normalizing flows}}
    \begin{itemize}
        \item Base density layers: 4 or 8
        \item Time-dependent NF layers: 4 or 8
    \end{itemize}
\end{itemize}

\end{document}